\documentclass[suppldata]{interact}

\usepackage{hyperref}
\usepackage{multirow}
\usepackage{epstopdf}% To incorporate .eps illustrations using PDFLaTeX, etc.
\usepackage[caption=false]{subfig}% Support for small, `sub' figures and tables
\usepackage[natbibapa,nodoi]{apacite}% Citation support using apacite.sty. Commands using natbib.sty MUST be deactivated first!
\theoremstyle{plain}% Theorem-like structures provided by amsthm.sty

\theoremstyle{definition}

\theoremstyle{remark}

\begin{document}

\articletype{}% Specify the article type or omit as appropriate

\title{Gender stereotypes in the mediated personalization of politics: Empirical evidence from a lexical, syntactic and sentiment analysis}

\author{
\name{Emanuele Brugnoli\thanks{CONTACT Emanuele Brugnoli. Email: brugnoliema@gmail.com}, Rosaria Simone and Marco Delmastro}
%\affil{}
}

\maketitle

\begin{abstract}
The media attention to the personal sphere of famous and important individuals has become a key element of the gender narrative. Here we combine lexical, syntactic and sentiment analysis to investigate the role of gender in the personalization of a wide range of political office holders in Italy during the period 2017-2020. On the basis of a score for words that is introduced to account for gender unbalance in both representative and news coverage, we show that the political personalization in Italy is more detrimental for women than men, with the persistence of entrenched stereotypes including a masculine connotation of leadership, the resulting women's unsuitability to hold political functions, and a greater deal of focus on their attractiveness and body parts. In addition, women politicians are covered with a more negative tone than their men counterpart when personal details are reported. Further, the major contribution to the observed gender differences comes from online news rather than print news, suggesting that the expression of certain stereotypes may be better conveyed when click baiting and personal targeting have a major impact.
\end{abstract}


\section{Introduction}

Gender stereotypes are widespread and display structural effects even in more egalitarian and more developed countries \citep{Bre2020}. They may contribute to gender disparities in the pursuit of societally important fields \citep{Mas2021}. Their origin can be linked to the social and cultural environment, and the educational system \citep{Car2019}. In this context, information plays a fundamental role in generating, disseminating and reinforcing gender stereotypes. For instance, the media image of politics as a masculine realm \citep{Bau2015} can depress the political ambitions of young women and discourage political elites from selecting women \citep{Van2020}.

The current media communication is even more characterized by sensation and entertainment \citep{Orn2004} and the phenomenon of personalization become a fundamental concept in the discussion on how political news evolves through time \citep{Lan2013}. At a coarse level, one speaks of personalization for referring to a focus on individual politicians rather than on the institutions they represent. More subtly, personalization implies a shift in media focus from the politician as a public office holder to the politician as a private individual. In the former case it is labelled as ``individualization'', in the latter case as ``privatization'' \citep{Van2012}. In this realm, gender stereotypes can be translated into the association of women politicians with private life \citep{One2016}, physical beauty \citep{Con2015} and supporting roles \citep{Koe2011}.
 
The aim of this article is therefore to assess the presence of gender stereotypes in the news through an innovative data driven approach based on lexical, syntactic and sentiment analysis to compare the media attention addressed to men and women politicians in a statistically sound way. With some more details, first we compare the Italian media coverage of men and women politicians for a wide and differentiated number of public roles: minsters, undersecretaries of state, governors and mayors of cities with more than sixty thousand inhabitants. So, the analyzed universe of politicians is large and representative of all political parties in Italy. In addition, the analysis covers a four-year timeframe (2017-2020) that comprises two changes of government, a constitutional referendum, a general election, several both local and regional elections, and the occurrence of considerable events (e.g., the outbreak of the Covid-19 pandemic and the related social and economic effects and policies)\footnote{Most existing studies concern a single context, and this could lead to stronger gender bias in reporting. For instance, the political actors under scrutiny are most powerful offices rather than representatives at local level \citep{Atk2008}, the focus is most solely on electoral campaigns and rarely even routine time \citep{Aal2020,Ger2009}, the majority of the extant work is conducted in the United States and less work is done in multi-party systems \citep{Van2020}.}. Second, we analyze the universe of all the articles (i.e., more than 1.8 million news items) reported in all national (and multiregional) newspapers and online news outlets which are related to the selected politicians. It is worth mentioning that these news sources reach the vast majority of citizens who get informed\footnote{Note that most of existing studies rely instead on the content analysis of a relatively small amount of articles, which in turn allows to manually identify the presence of personalizing elements and assign a polarity orientation \citep{Van2012,Tri2013,Wag2019}.}. Third, we define a robust methodology to identify and then statistically analyze the lexical, syntactic and sentiment gender patterns of news production. Namely, we build a lexicon of words which account for personal details (i.e., physical traits, moral attitudes, and economic and financial background) and are attributed to the political actors under scrutiny by analyzing the syntactic dependencies of the politician-word pair in the sentences where they are both mentioned. In addition, for each of these terms we determine its semantic orientation in the political domain.
 
The proposal is robust with respect to the structural gender unbalance in both representative and coverage: specifically, the exploratory data analysis relies on the definition of a coverage index adjusted for gender bias, that allows to safely measure the diversity in incidence, stratified for word category, and identify gender-distinctive words. Quantile regression is then applied to jittered sentiment scores to assess the extent to which differences related to the gender of the politician and to the type of news source (print or online) are significant and relevant.

The findings highlight the existence of persistent, entrenched gender stereotypes, especially when going online (i.e., in online news outlets compared to traditional newspapers). Women politicians receive more focus on the privatization dimension than men (physical and socio-economic characteristics), whereas the coverage of their men colleagues is higher on the individualization dimension (moral and behavioral characteristics). In particular, men are depicted as powerful, active and violent, while women are told as not fit to hold a public office, concentrating a greater deal of focus on their attractiveness and body parts. Finally, we find that women politicians are depicted with a more negative tone with respect to each of the analyzed categories.

\section{Materials and methods}\label{mam}

\subsection{The selection of news media sources}

To ensure the most representative picture of both traditional and new media, we considered a wide range of national and local newspapers and online news outlets that are active in Italy during the period January 2017 - November 2020 (see Supplemental online material for the complete list of sources). We selected all the major Italian newspapers which are the ones that still have a great influence on the political agenda \citep{Dra2014}. In 2020, the 83 considered newspapers reached 22 million Italians, i.e., 43\% of the population aged more than 18 (source: \href{https://www.gfk.com/home}{GfK Mediamonitor}). We also included as sources more than 250 online-only news outlets, that monthly reach 38 million Italians, i.e., 93\% of the total internet audience (source: \href{https://www.comscore.com/}{ComScore}). In sum, we considered the universe of online-only and traditional news sources covering a broad spectrum of points of views and reaching the great majority of Italian citizens who get informed.

\subsection{The proposed approach}

\begin{figure}[h]
	\centering
	\includegraphics[width=0.5\columnwidth]{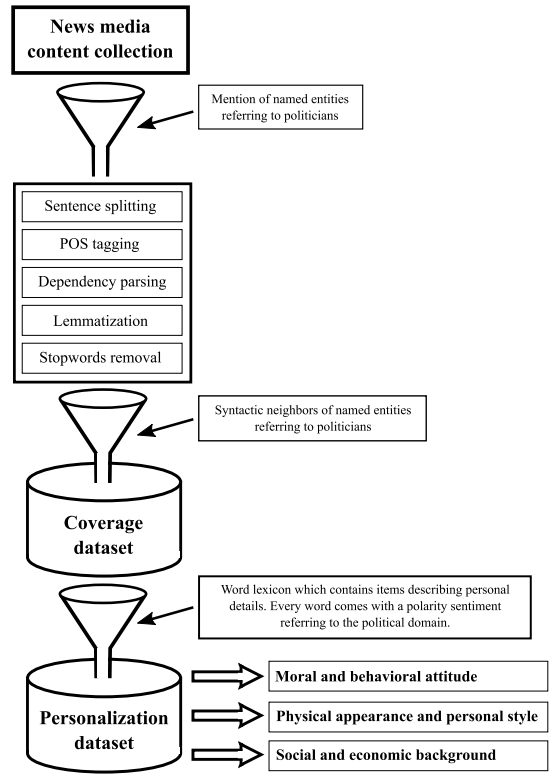}
	\caption{Overview of the proposed approach.}\label{fig:overview}
\end{figure}

Figure \ref{fig:overview} shows the architectural overview of our method.

The input to the system is a collection of news items filtered according to the occurrence of named entities referring to the political offices under scrutiny. The output of the system is an assessment of the personalized coverage and the corresponding sentiment concerning the politicians under investigation. All the procedural steps are illustrated in Supplemental online material. 

To identify the personalizing elements of the coverage, we construct a word lexicon based on several key indicators which are representative of the media coverage of personal details \citep{Van2012,Tri2013}. The lexicon is divided into three categories that aim at capturing the context of each word: i) moral and behavioral attitude; ii) physical appearance and personal style; and iii) social and economic background. For each of these terms we determine its semantic orientation in the political domain.

\subsection{Data collection}
To include offices at both local and national level, the target under scrutiny comprises all the Italian politicians serving as minsters, undersecretaries of state, governors and mayors of cities with more than sixty thousand inhabitants. Presidents of the Republic and Prime Ministers are both excluded from the analysis, since no woman has ever occupied such roles in Italy. Data have been gathered by means of a media monitoring platform developed by the IT company \href{https://www.volocom.it/?lang=en}{Volocom Technology}. The exact breakdown of both the coverage dataset ($\mathcal{D}_c$) and the personalization datasets ($\mathcal{D}_p$) is presented in Table \ref{tab:coverage_dataset_breakdown}.
\begin{table}[h]
	\centering
	\begin{tabular}{lrrrr}
		\toprule
		& \multicolumn{2}{r}{\textbf{Coverage dataset}} & \multicolumn{2}{r}{\textbf{Personalization dataset}} \\
		& F & M & F & M\\
		\midrule
		Politicians & 57 & 213 & 56 & 211\\
		Contents & 328,842 & 1,519,115 & 19,185 & 82,429\\
		Sentences & 689,574 & 3,368,608 & 21,599 & 97,589\\
		Words & 929,160 & 5,075,651 & 23,875 & 110,765\\
		Distinct words & 17,722 & 36,238 & 1,357 & 1,793\\
		\bottomrule
	\end{tabular}
	\caption{Breakdown of both the coverage and personalization datasets divided by gender.}
	\label{tab:coverage_dataset_breakdown}
\end{table}
The reported values concern the number of political offices under scrutiny; the number of media contents with the mention of at least one of such politicians; the related sentences containing such mentions; the number of words and unique words, respectively, contained in the syntactic neighborhood of the named entities mentioned. Note that the term ``word'' is used for referring to its base form. Moreover, albeit we aim to refer to sentences as coded units to analyze, for the sake of simplicity we consider words instead. Indeed, the syntactic neighborhood of the named entity mentioned consists of a single lexicon word in almost all the sentences in $\mathcal{D}_p$ (see Supplemental online material for further details).

\subsection{POS tagging and dependency parsing}
Part-of-speech (POS) information can be considered the first step in semantic disambiguation \citep{Wil1998} and sentiment analysis \citep{Pan2008}. Adjectives are indeed considered the primary source of subjective content \citep{Hat2000,Whi2005,Yu2003} as well as the gauge of personality features of politicians \citep{Cap2007,Cap2006,Cap2008,Sim1986}. Nevertheless, this does not imply that other parts of speech cannot be used for referring to personal details. We argue that nouns (e.g., \textit{skirt, son, hair}) as well as verbs (e.g., \textit{love, wear, tease}) can also be strong indicators for personalization \citep{Fas2016}, then we also consider them as sources of subjective content to analyze.

For identifying the words in a sentence which are actually attributed to a given target, linear $n$-grams in the sense of adjacent strings of tokens, parts of speech, etc. could be not satisfactory (see Supplemental online material for a detailed discussion). To overcome this problem we rely on the syntactic $n$-grams methodology, i.e. instead of following the sequential order in the sentence, the linguistic pattern of the words is based on their respective position in the syntactic parse tree. We argue that the words which appear nearby a named entity in the dependency tree are more likely candidates for personalizing expressions than those farther by. Through the SpaCy linguistic parser \citep{Hon2020} trained on a corpus of annotated news media texts in Italian \cite{Bos2014,Bos2013}, we first split the text of each media content into sentences, then we produce the POS tag for each word and the syntactic tree of each sentence.

\subsection{A lexicon of semantic-oriented words describing personal details in the political domain}
To the best of our knowledge, there are no publicly available lexical resources in Italian language which are designed to account for the personalization phenomenon in the political domain. Hence, we decide to create a manual lexicon, starting from a selection of suitable words (1,249 unique lemmas) extracted from a preexisting lexicon of hate words \citep{Bas2018}. As a second step, we expand the lexicon by systematically investigating key indicators of personalized news coverage, i.e., personality traits and behavioral attitude, coverage of the family, past life and upbringing, leisure time, love life, age, appearance, social background and economic opportunities \citep{Van2012,Tri2013}. The third step consists of identifying any further personalizing word in the coverage dataset, and then ensures an exhaustive inventory of all the personalizing terms occurring in the news media contents under investigation. The final lexicon is composed of 3,303 words divided in 2,125 adjectives, 1,084 nouns and 94 verbs.

Once the lexicon is complete, we deal with the semantic orientation of the single words. To this aim, we hire five annotators for manually assigning to each word one of the following sentiment scores: -1, 0 and 1 for negative, neutral and positive meanings, respectively. To summarize the semantic orientation of a single word in our lexicon, we assign it the average value of the five scores received during the annotation process. Hence, the aggregate sentiment score assigned to a lexicon word can be one of the eleven terms of the sequence $\big(\frac{k-5}{5}\big)_{k=0}^{10}$. The resulting values are then grouped into ordinal categories: negative (strong and weakly), neutral, positive (weakly and strong). See Supplemental online material also for downloading the resource.

\subsection{An index reporting gender homogeneity in coverage, after adjusting for coverage bias}
Since the political offices in Italy are mainly coupled with men candidates, this naturally implies that the whole women representative receives less media coverage than the men counterpart. Therefore, to compare the words' coverage per women and men, respectively, we need to define a gendered score for each word that takes into account the women-men unbalance concerning both the number of politicians and gender-coverage. Following the methodology reported in Supplemental online material, the score of a word $w$ is measured by the coverage bias index $I$ given by the normalized difference between the (adjusted) incidence rate $\tilde{t}_F(w)$ associating the word with women and the (adjusted) incidence rate $\tilde{t}_M(w)$ associating the word with men (See Supplemental online material for details), that is:
\begin{equation}
	\label{index1}
	I(w) = \frac{\tilde{t}_F(w)-\tilde{t}_M(w)}{\tilde{t}_F(w)+\tilde{t}_M(w)}, \qquad I(w) \in [-1,1].
\end{equation}
It is straightforward to notice that $I(w)=1$ if and only if $w$ is used exclusively for women politicians, whereas $I(w)=-1$ if and only if $w$ is used for their men colleagues only. See Supplemental online material for the definition of the adjusted incidence rate and for a discussion on the reliability of the coverage bias index I under different scenarios.

\subsection{Dissimilarity of word frequency distributions}
Aside from studying the distribution of the coverage bias index $I$, we also pursue an analysis of the words' frequency distributions with the goal of determining possible gender-distinctive words. To this aim, we borrow the rationale of Leti diversity index \citep{Let1983} and we define an index of dissimilarity between women and men representations as follows:
\begin{equation}\label{diss1}
	Diss = \dfrac{c_F\cdot c_M}{c_F+c_M}\sum_{w \in \mathcal{D}_c}|\tilde{t}_F(w) - \tilde{t}_M(w)|,  \qquad Diss \in [0,1].
\end{equation}
where $c_F$ and $c_M$ are the correction factors defined to adjust the aforementioned incidence rates and thus make them comparable in view of the strong unbalance of the dataset (See Supplemental online material for details). Next, we compute the \textit{leave-one-out} dissimilarity to identify gender-distinctive personalizing words. Thus, for each word $w^{\ast}\in \mathcal{D}_c$ we compute the dissimilarity between men and women frequency distributions obtained after omitting $w^{\ast}$, namely:
\begin{equation}\label{diss2}
	Diss_{(-w^{\ast})} = \dfrac{c_F^{\ast}\cdot c_M^{\ast}}{c_F^{\ast}+c_M^{\ast}}\sum_{\substack{w \in \mathcal{D}_c\\w \neq w^{\ast}}}|\tilde{t}_F^{\ast}(w) - \tilde{t}_M^{\ast}(w)|, \qquad Diss_{(-w^{\ast})} \in [0,1],
\end{equation}
where the superscript $^{\ast}$ means that correction factors and adjusted incidence rates are calculated on $\mathcal{D}_c\setminus\{w^{\ast}\}$. Finally, we identify as gender-distinctive those words $w^{\ast}$ such that $Diss_{(-w^{\ast})}<Diss$, namely those words whose omission from $\mathcal{D}_c$ contributes to reduce the dissimilarity of words coverage between gender. In particular, a word $w^{\ast}$ such that $Diss_{(-w^{\ast})}<Diss$ is considered men-distinctive if $\tilde{t}_M(w)>\tilde{t}_F(w)$ and women-distinctive otherwise.

\section{Results}

\subsection{Gender gaps in the mediated personalization of politics}
\begin{figure}[h]
	\centering
	\includegraphics[width=\columnwidth]{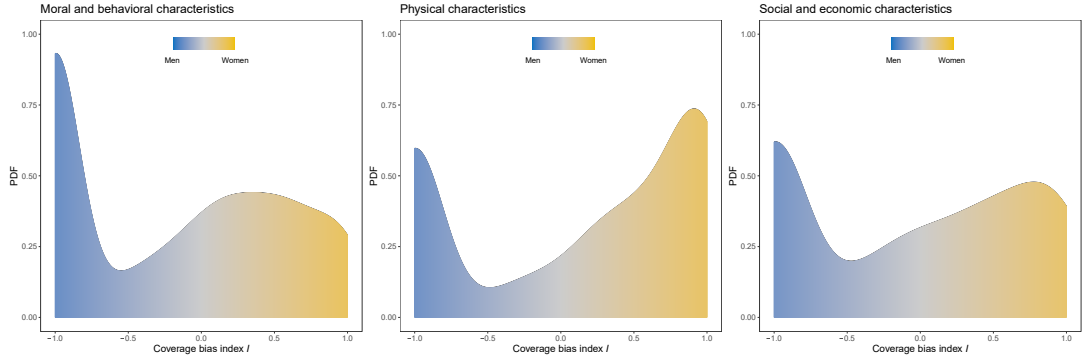}
	\caption{Empirical PDF of the coverage bias index $I$ evaluated over the personalizing wording with regard to the political actors under scrutiny. Given a word $w$, $I(w)$ represents the normalized difference in using $w$ for women and men representations.}\label{fig:pdf}
\end{figure}

Figure \ref{fig:pdf} shows the Probability Density Function (PDF) of the coverage bias index $I$ defined in \eqref{index1} over the personalizing wording with regard to the political actors under scrutiny, conditional to each analyzed category.

Moreover, Table \ref{tab:index_summary} reports some descriptive statistics for the coverage bias index $I$ for the total counts per words category.
\begin{table}[h]
	\centering
	\begin{tabular}{rrrrrrrrr}
		\toprule
		& $\mu$ & $\gamma_3$ & $D_5$ & $Q_3$ & $D_9$ & IQR\\
		\midrule
		\textbf{Moral and behavioral} & -0.185 & 0.142 & -0.122 & 0.454 & 0.828 & 1.454\\
		\textbf{Physical} & 0.134 & -0.419 & 0.391 & 0.931 & 1.000 & 1.931\\
		\textbf{Social and economic} & -0.074 & -0.029 & 0.005 & 0.647 & 1.000 & 1.647\\
		\bottomrule
	\end{tabular}
	\caption{Summary descriptive statistics for the coverage bias index $I$ - words count per category of personalization: average $\mu$; Fisher index of asymmetry $\gamma_3$; median $D_5$; third quartile $Q_3$; ninth decile $D_9$; inter-quartile range (IQR).}\label{tab:index_summary}
\end{table}

Assuming that gender balance would correspond to a symmetric distribution with mean at $I=0$, evidence is found that political coverage is biased in favour of men with respect to moral and behavioral characteristics. On the contrary, the coverage bias index presents a strong negative skewness for physical characteristics, which along with a positive average, indicate that political journalism towards women focuses a strong amount of attention to physical characteristics. It should be noted that this result is also confirmed by the time analysis that shows a persistent and structural difference typical of entrenched stereotypes (see Supplemental online material for details).

\subsection{The role of gender in the quality of coverage and in the sentiment expressed through personalization}
Besides wondering whether women politicians receive more media attention on personal details, we also account for gender differences in the ways those details are reported. To this aim, among the gender-distinctive personalizing words of each category, we select those words $w^{\ast}$ for which the difference $Diss_{(-w^{\ast})}<Diss$ is large enough (see \eqref{diss1} and \eqref{diss2}). This filtering returns men politicians stereotypically depicted as powerful, active and violent. On the contrary, women are strongly perceived as not fit to hold public office. It is also interesting to note that all the words referring to parenting are unbalanced towards women, as if to stress the role played by powerful parents in the political careers of their daughters. With respect to physical characteristics instead, men politicians are mainly depicted with reference to size while women receive a greater deal of focus on their attractiveness and body parts (see Supplemental online material for details).

The lexicon words used to identify the personalized items of media coverage are also annotated with the semantic orientation assigned by five annotators hired to this aim. The reliability of the annotation process turns out to be fairly high, as gauged by the Krippendorff's $\alpha=0.712$. Then, we rely on the average values of the single sentiment scores assigned to each personalizing word to evaluate the gender differences in the sentiment expressed through personalization.
Table \ref{tab:sentiment} shows, for each analyzed facet of the personalization phenomenon, the fraction of negative, neutral and positive wording with regard to the women and men representations. To be thorough, we also report the distribution of the lexicon words over the sentiment categories.
\begin{table}[h]
	\centering\small
	\begin{tabular}{llrrrrr}
		\toprule
		& & \multicolumn{2}{c}{\textbf{Negative}} & \textbf{Neutral} & \multicolumn{2}{c}{\textbf{Positive}} \\
		& & strong & weakly & & weakly & strong \\
		\midrule
		\multirow{3}{*}{\textbf{Moral and behavioral}} & \textit{Lexicon} & $\mathit{51.09\%}$ & $\mathit{15.86\%}$ & $\mathit{9.25\%}$ & $\mathit{7.93\%}$ & $\mathit{15.87\%}$ \\
		& Men & $23.24\%$ & $18.90\%$ & $\mathbf{19.24\%}$ & $\mathbf{22.35\%}$ & $\mathbf{16.28\%}$ \\
		& Women & $\mathbf{28.20\%}$ & $\mathbf{19.80\%}$ & $16.14\%$ & $19.77\%$ & $16.19\%$ \\
		\midrule
		\multirow{3}{*}{\textbf{Physical}} & \textit{Lexicon} & $\mathit{29.41\%}$ & $\mathit{23.26\%}$ & $\mathit{28.18\%}$ & $\mathit{13.54\%}$ & $\mathit{5.61\%}$ \\
		& Men & $13.39\%$ & $13.82\%$ & $\mathbf{46.30\%}$ & $\mathbf{22.79\%}$ & $3.70\%$ \\
		& Women & $\mathbf{15.28\%}$ & $\mathbf{17.49\%}$ & $40.03\%$ & $21.16\%$ & $\mathbf{6.04\%}$ \\
		\midrule
		\multirow{3}{*}{\textbf{Social and economic}} & \textit{Lexicon} & $\mathit{41.02\%}$ & $\mathit{17.22\%}$ & $\mathit{26.74\%}$ & $\mathit{13.19\%}$ & $\mathit{1.83\%}$ \\
		& Men & $\mathbf{5.32\%}$ & $14.91\%$ & $\mathbf{54.54\%}$ & $23.23\%$ & $\mathbf{2.00\%}$ \\
		& Women & $3.08\%$ & $\mathbf{15.30\%}$ & $48.56\%$ & $\mathbf{31.97\%}$ & $1.09\%$ \\
		\bottomrule
	\end{tabular}
	\caption{Fraction of negative, neutral and positive wording with regard to lexicon, men and women representations, respectively, for each analyzed facet of personalization.}
	\label{tab:sentiment}
\end{table}

As highlighted with bold font, the negative tone is always greater (in percent) on women politicians than on their men counterparts, with the only exception of strong negative descriptions concerning the socio-economic category. A reverse trend concerns instead neutral and uplifting portrayals. 

\subsection{Print news versus online news: the personalization phenomenon as a function of the type of media source}
Compared to print newspapers, online news outlets have a number of characteristics that can affect the personalization phenomenon and widen the differences between women and men representations. Indeed, online-only news outlets are presumed to be influenced more strongly by personalized algorithms, click baiting phenomenon, and individual comments on news stories \citep{Sko2014}. To check this hypothesis we consider the frequency distribution of words count per gender conditional to both dataset (coverage and personalization) and source type (traditional newspapers and online news outlets). The $\chi^2$ test of independence for both these contingency tables is highly significant, indicating a strong association between gendered coverage (personalization) and source type. Specifically, observed coverage (personalization) provided by online sources is higher than expected under the assumption of independence for women, whereas it is lower than expected for men. The converse is true for traditional sources: observed coverage (personalization) for women is lower than expected if no association were present, whereas it is higher for men (see Table S4 in Supplemental online material).  
The empirical distribution of the coverage-bias index $I$ given source type is substantially similar to the PDF of Figure \ref{fig:pdf}, with respect to any of the considered personalization categories, both for traditional newspapers and online news outlets. Hence, political coverage results biased in favour of men with respect to moral and behavioral characteristics, whereas it results biased towards women with regard to physical characteristics, both for traditional newspapers and online news outlets. The coverage bias density distribution for socio-economic characteristics, instead, is more heterogeneous.
Concerning the tone of personalized coverage, we estimate a quantile regression model based on the observations of the personalization dataset (conditional to each analyzed category) for the (jittered) sentiment score distribution ($Y_i$) as a function of dummy variables for Gender, Source type, and their interaction: 
\begin{equation}
	\label{quantilereg}
	\text{Quantile}(Y_i)= \beta_0+\beta_1\text{Gender}_i+\beta_2\text{Source}_i+\beta_3\text{Gender}_i\cdot\text{Source}_i.
\end{equation}
Table \ref{tab:regression} reports the estimated conditional quantiles for each of the 12 groups identified by cross-classifying gender, source type and word categories. Specifically, the quantile regression was meant to test: i) if strong and weak negative tones (in terms of first decile $D_1$ and first quartile $Q_1$, resp.), neutral tone (in terms of median $D_5$), and weak and positive tones (in terms of third quartile $Q_3$ and ninth decile $D_9$) are addressed to women and men in a significantly different way; ii) if this circumstance depends in turn on the source type; and iii) the extent by which gender differences vary from tradition to online sources.
\begin{table}[h]
	\centering\footnotesize
	\begin{tabular}{rcrrrrrr}
		\toprule
		Category & Gender & Source type & $D_1$ & $Q_1$ & $D_5$ & $Q_3$ & $D_9$\\
		\midrule
		\multirow{4}{*}{\textbf{Moral and behavioral}} & \multirow{2}{*}{F} & Online & $-1.000$ & $-0.801$ & $-0.208$ & $0.593$ & $0.813$ \\
		& & Traditional & $-0.999$ & $-0.795$ & $-0.203$ & $0.596$ & $0.813$ \\
		& \multirow{2}{*}{M} & Online & $-0.999$ & $-0.792$ & $-0.195$ & $0.598$ & $0.977$ \\
		& & Traditional & $-0.995$ & $-0.606$ & $-0.007$ & $0.599$ & $0.810$ \\
		\midrule
		\multirow{4}{*}{\textbf{Physical}} & \multirow{2}{*}{F} & Online & $-0.997$ & $-0.598$ & $-0.002$ & $0.393$ & $0.600$ \\
		& & Traditional & $-0.815$ & $-0.589$ & $-0.002$ & $0.388$ & $0.598$ \\
		& \multirow{2}{*}{M} & Online & $-0.806$ & $-0.394$ & $0.001$ & $0.391$ & $0.597$ \\
		& & Traditional & $-0.799$ & $-0.400$ & $0.000$ & $0.384$ & $0.592$ \\
		\midrule
		\multirow{4}{*}{\textbf{Social and economic}} & \multirow{2}{*}{F} & Online & $-0.595$ & $-0.193$ & $0.004$ & $0.398$ & $0.596$ \\
		& & Traditional & $-0.413$ & $-0.015$ & $0.007$ & $0.401$ & $0.599$ \\
		& \multirow{2}{*}{M} & Online & $-0.592$ & $-0.021$ & $0.002$ & $0.212$ & $0.411$ \\
		& & Traditional & $-0.594$ & $-0.194$ & $-0.002$ & $0.385$ & $0.590$ \\
		\bottomrule
	\end{tabular}
	\caption{Estimated (conditional) quantiles from regression model \eqref{quantilereg}.}
	\label{tab:regression}
\end{table}

Hereafter, we comment only on the significant results: with the only exception of socio-economic facet for men politicians, negative sentiment results stronger for online news outlets than it is for traditional newspapers. This is especially true for physical and socio-economic features of the women representative, and moral-behavioral details of the men counterpart. Moreover, with the only exception of traditional coverage on socio-economic details, negative sentiment is stronger for women than it is for men. This is true to a greater extent for online coverage on physical characteristics.

\section{Discussion}
This paper provides robust evidence on the presence of different and stereotyped narratives of news media when dealing with the gender of the politicians. The space of our investigation is represented by all the articles produced by almost the entire universe of Italian traditional newspapers and online news outlets over the four-year period 2017-2020. Our method relies on a hybrid approach combining lexical, syntactic and sentiment analysis. Namely, we build a lexicon of words which account for personal details and are attributed to the political actors under scrutiny by analyzing the syntactic dependencies of the politician-word pair in the sentences where they are both mentioned. In addition, for each of these terms we determine its semantic orientation in the political domain. Further, since the political offices in Italy are mainly coupled with men candidates, we introduce on a statistical index which assigns a gender bias coverage score to each word by taking into account the women-men unbalance concerning both the number of politicians and coverage.
Our findings show that personalization in Italy is still a gendered phenomenon, with women politicians typically receiving more mentions (in percent) to their private, i.e., physical and socio-economic characteristics, throughout the period under investigation.
Moreover, an assessment of the differences in the ways politicians are discussed reveals that stereotypically men are depicted as powerful, active and violent, whereas women are strongly perceived as not fit to hold a public office. In addition, with respect to physical appearance, women politicians receive a greater deal of focus on their attractiveness and their body parts.
Finally, by investigating the personalization phenomenon as a function of the type of source, we find that the major contribution to the personalized overrepresentation and more negative sentiment concerning women politicians comes from online news outlets rather than traditional newspapers, suggesting that the expression of certain stereotypes may be better conveyed when personalized algorithms and click baiting logics have a major impact.

\section*{Funding}
This work was partially supported by the European Union’s Rights, Equality and Citizenship Programme (2014-2020) under Grant n. 875263.

\bibliographystyle{apacite}
\bibliography{interactapasample}

\begin{thebibliography}{}

\bibitem [\protect \citeauthoryear {%
Aaldering%
\ \BBA {} Van Der~Pas%
}{%
Aaldering%
\ \BBA {} Van Der~Pas%
}{%
{\protect \APACyear {2020}}%
}]{%
Aal2020}
\APACinsertmetastar {%
Aal2020}%
\begin{APACrefauthors}%
Aaldering, L.%
\BCBT {}\ \BBA {} Van Der~Pas, D\BPBI J.%
\end{APACrefauthors}%
\unskip\
\newblock
\APACrefYearMonthDay{2020}{}{}.
\newblock
{\BBOQ}\APACrefatitle {Political Leadership in the Media: Gender Bias in Leader
  Stereotypes during Campaign and Routine Times} {Political leadership in the
  media: Gender bias in leader stereotypes during campaign and routine
  times}.{\BBCQ}
\newblock
\APACjournalVolNumPages{British Journal of Political
  Science}{50}{3}{911–931}.
\newblock
\begin{APACrefDOI} \doi{https://doi.org/10.1017/S0007123417000795}
  \end{APACrefDOI}
\PrintBackRefs{\CurrentBib}

\bibitem [\protect \citeauthoryear {%
Atkeson%
\ \BBA {} Krebs%
}{%
Atkeson%
\ \BBA {} Krebs%
}{%
{\protect \APACyear {2008}}%
}]{%
Atk2008}
\APACinsertmetastar {%
Atk2008}%
\begin{APACrefauthors}%
Atkeson, L\BPBI R.%
\BCBT {}\ \BBA {} Krebs, T\BPBI B.%
\end{APACrefauthors}%
\unskip\
\newblock
\APACrefYearMonthDay{2008}{}{}.
\newblock
{\BBOQ}\APACrefatitle {Press Coverage of Mayoral Candidates: The Role of Gender
  in News Reporting and Campaign Issue Speech} {Press coverage of mayoral
  candidates: The role of gender in news reporting and campaign issue
  speech}.{\BBCQ}
\newblock
\APACjournalVolNumPages{Political Research Quarterly}{61}{2}{239-252}.
\newblock
\begin{APACrefDOI} \doi{https://doi.org/10.1177/1065912907308098}
  \end{APACrefDOI}
\PrintBackRefs{\CurrentBib}

\bibitem [\protect \citeauthoryear {%
Bassignana%
, Basile%
\BCBL {}\ \BBA {} Patti%
}{%
Bassignana%
\ \protect \BOthers {.}}{%
{\protect \APACyear {2018}}%
}]{%
Bas2018}
\APACinsertmetastar {%
Bas2018}%
\begin{APACrefauthors}%
Bassignana, E.%
, Basile, V.%
\BCBL {}\ \BBA {} Patti, V.%
\end{APACrefauthors}%
\unskip\
\newblock
\APACrefYearMonthDay{2018}{}{}.
\newblock
{\BBOQ}\APACrefatitle {Hurtlex: A multilingual lexicon of words to hurt}
  {Hurtlex: A multilingual lexicon of words to hurt}.{\BBCQ}
\newblock
\BIn{} \APACrefbtitle {5th Italian Conference on Computational Linguistics,
  CLiC-it 2018} {5th italian conference on computational linguistics, clic-it
  2018}\ (\BVOL\ 2253, \BPG~1-6).
\PrintBackRefs{\CurrentBib}

\bibitem [\protect \citeauthoryear {%
Bauer%
}{%
Bauer%
}{%
{\protect \APACyear {2015}}%
}]{%
Bau2015}
\APACinsertmetastar {%
Bau2015}%
\begin{APACrefauthors}%
Bauer, N\BPBI M.%
\end{APACrefauthors}%
\unskip\
\newblock
\APACrefYearMonthDay{2015}{}{}.
\newblock
{\BBOQ}\APACrefatitle {Emotional, Sensitive, and Unfit for Office? {G}ender
  Stereotype Activation and Support Female Candidates} {Emotional, sensitive,
  and unfit for office? {G}ender stereotype activation and support female
  candidates}.{\BBCQ}
\newblock
\APACjournalVolNumPages{Political Psychology}{36}{6}{691-708}.
\newblock
\begin{APACrefDOI} \doi{https://doi.org/10.1111/pops.12186} \end{APACrefDOI}
\PrintBackRefs{\CurrentBib}

\bibitem [\protect \citeauthoryear {%
Bosco%
, Dell’Orletta%
, Montemagni%
, Sanguinetti%
\BCBL {}\ \BBA {} Simi%
}{%
Bosco%
\ \protect \BOthers {.}}{%
{\protect \APACyear {2014}}%
}]{%
Bos2014}
\APACinsertmetastar {%
Bos2014}%
\begin{APACrefauthors}%
Bosco, C.%
, Dell’Orletta, F.%
, Montemagni, S.%
, Sanguinetti, M.%
\BCBL {}\ \BBA {} Simi, M.%
\end{APACrefauthors}%
\unskip\
\newblock
\APACrefYearMonthDay{2014}{dec}{}.
\newblock
{\BBOQ}\APACrefatitle {The {E}valita 2014 Dependency Parsing Task} {The
  {E}valita 2014 dependency parsing task}.{\BBCQ}
\newblock
\BIn{} \APACrefbtitle {Proceedings of Evalita ’14, Evaluation of NLP and
  Speech Tools for Italian} {Proceedings of evalita ’14, evaluation of nlp
  and speech tools for italian}\ (\BPG~1-8).
\newblock
\APACaddressPublisher{Pisa, Italy}{Association for Computational Linguistics}.
\PrintBackRefs{\CurrentBib}

\bibitem [\protect \citeauthoryear {%
Bosco%
, Montemagni%
\BCBL {}\ \BBA {} Simi%
}{%
Bosco%
\ \protect \BOthers {.}}{%
{\protect \APACyear {2013}}%
}]{%
Bos2013}
\APACinsertmetastar {%
Bos2013}%
\begin{APACrefauthors}%
Bosco, C.%
, Montemagni, S.%
\BCBL {}\ \BBA {} Simi, M.%
\end{APACrefauthors}%
\unskip\
\newblock
\APACrefYearMonthDay{2013}{aug}{}.
\newblock
{\BBOQ}\APACrefatitle {Converting {I}talian Treebanks: Towards an {I}talian
  {S}tanford Dependency Treebank} {Converting {I}talian treebanks: Towards an
  {I}talian {S}tanford dependency treebank}.{\BBCQ}
\newblock
\BIn{} \APACrefbtitle {Proceedings of the 7th Linguistic Annotation Workshop
  and Interoperability with Discourse} {Proceedings of the 7th linguistic
  annotation workshop and interoperability with discourse}\ (\BPG~61-69).
\newblock
\APACaddressPublisher{Sofia, Bulgaria}{Association for Computational
  Linguistics}.
\PrintBackRefs{\CurrentBib}

\bibitem [\protect \citeauthoryear {%
Breda%
, Jouini%
, Napp%
\BCBL {}\ \BBA {} Thebault%
}{%
Breda%
\ \protect \BOthers {.}}{%
{\protect \APACyear {2020}}%
}]{%
Bre2020}
\APACinsertmetastar {%
Bre2020}%
\begin{APACrefauthors}%
Breda, T.%
, Jouini, E.%
, Napp, C.%
\BCBL {}\ \BBA {} Thebault, G.%
\end{APACrefauthors}%
\unskip\
\newblock
\APACrefYearMonthDay{2020}{}{}.
\newblock
{\BBOQ}\APACrefatitle {Gender stereotypes can explain the gender-equality
  paradox} {Gender stereotypes can explain the gender-equality paradox}.{\BBCQ}
\newblock
\APACjournalVolNumPages{Proceedings of the National Academy of
  Sciences}{117}{49}{31063-31069}.
\newblock
\begin{APACrefDOI} \doi{https://doi.org/10.1073/pnas.2008704117}
  \end{APACrefDOI}
\PrintBackRefs{\CurrentBib}

\bibitem [\protect \citeauthoryear {%
Caprara%
}{%
Caprara%
}{%
{\protect \APACyear {2007}}%
}]{%
Cap2007}
\APACinsertmetastar {%
Cap2007}%
\begin{APACrefauthors}%
Caprara, G\BPBI V.%
\end{APACrefauthors}%
\unskip\
\newblock
\APACrefYearMonthDay{2007}{}{}.
\newblock
{\BBOQ}\APACrefatitle {The Personalization of Modern Politics} {The
  personalization of modern politics}.{\BBCQ}
\newblock
\APACjournalVolNumPages{European Review}{15}{2}{151–164}.
\newblock
\begin{APACrefDOI} \doi{https://doi.org/10.1017/S1062798707000178}
  \end{APACrefDOI}
\PrintBackRefs{\CurrentBib}

\bibitem [\protect \citeauthoryear {%
Caprara%
, Schwartz%
, Capanna%
, Vecchione%
\BCBL {}\ \BBA {} Barbaranelli%
}{%
Caprara%
\ \protect \BOthers {.}}{%
{\protect \APACyear {2006}}%
}]{%
Cap2006}
\APACinsertmetastar {%
Cap2006}%
\begin{APACrefauthors}%
Caprara, G\BPBI V.%
, Schwartz, S.%
, Capanna, C.%
, Vecchione, M.%
\BCBL {}\ \BBA {} Barbaranelli, C.%
\end{APACrefauthors}%
\unskip\
\newblock
\APACrefYearMonthDay{2006}{}{}.
\newblock
{\BBOQ}\APACrefatitle {Personality and Politics: Values, Traits, and Political
  Choice} {Personality and politics: Values, traits, and political
  choice}.{\BBCQ}
\newblock
\APACjournalVolNumPages{Political Psychology}{27}{1}{1-28}.
\newblock
\begin{APACrefDOI} \doi{https://doi.org/10.1111/j.1467-9221.2006.00447.x}
  \end{APACrefDOI}
\PrintBackRefs{\CurrentBib}

\bibitem [\protect \citeauthoryear {%
Caprara%
, Schwartz%
, Vecchione%
\BCBL {}\ \BBA {} Barbaranelli%
}{%
Caprara%
\ \protect \BOthers {.}}{%
{\protect \APACyear {2008}}%
}]{%
Cap2008}
\APACinsertmetastar {%
Cap2008}%
\begin{APACrefauthors}%
Caprara, G\BPBI V.%
, Schwartz, S\BPBI H.%
, Vecchione, M.%
\BCBL {}\ \BBA {} Barbaranelli, C.%
\end{APACrefauthors}%
\unskip\
\newblock
\APACrefYearMonthDay{2008}{}{}.
\newblock
{\BBOQ}\APACrefatitle {The personalization of politics: Lessons from the
  {I}talian case} {The personalization of politics: Lessons from the {I}talian
  case}.{\BBCQ}
\newblock
\APACjournalVolNumPages{European Psychologist}{13}{3}{157–172}.
\newblock
\begin{APACrefDOI} \doi{https://doi.org/10.1027/1016-9040.13.3.157}
  \end{APACrefDOI}
\PrintBackRefs{\CurrentBib}

\bibitem [\protect \citeauthoryear {%
Carlana%
}{%
Carlana%
}{%
{\protect \APACyear {2019}}%
}]{%
Car2019}
\APACinsertmetastar {%
Car2019}%
\begin{APACrefauthors}%
Carlana, M.%
\end{APACrefauthors}%
\unskip\
\newblock
\APACrefYearMonthDay{2019}{03}{}.
\newblock
{\BBOQ}\APACrefatitle {{Implicit Stereotypes: Evidence from Teachers’ Gender
  Bias}} {{Implicit Stereotypes: Evidence from Teachers’ Gender
  Bias}}.{\BBCQ}
\newblock
\APACjournalVolNumPages{The Quarterly Journal of Economics}{134}{3}{1163-1224}.
\newblock
\begin{APACrefDOI} \doi{https://doi.org/10.1093/qje/qjz008} \end{APACrefDOI}
\PrintBackRefs{\CurrentBib}

\bibitem [\protect \citeauthoryear {%
Conroy%
, Oliver%
, Breckenridge-Jackson%
\BCBL {}\ \BBA {} Heldman%
}{%
Conroy%
\ \protect \BOthers {.}}{%
{\protect \APACyear {2015}}%
}]{%
Con2015}
\APACinsertmetastar {%
Con2015}%
\begin{APACrefauthors}%
Conroy, M.%
, Oliver, S.%
, Breckenridge-Jackson, I.%
\BCBL {}\ \BBA {} Heldman, C.%
\end{APACrefauthors}%
\unskip\
\newblock
\APACrefYearMonthDay{2015}{}{}.
\newblock
{\BBOQ}\APACrefatitle {From Ferraro to Palin: sexism in coverage of vice
  presidential candidates in old and new media} {From ferraro to palin: sexism
  in coverage of vice presidential candidates in old and new media}.{\BBCQ}
\newblock
\APACjournalVolNumPages{Politics, Groups, and Identities}{3}{4}{573-591}.
\newblock
\begin{APACrefDOI} \doi{https://doi.org/10.1080/21565503.2015.1050412}
  \end{APACrefDOI}
\PrintBackRefs{\CurrentBib}

\bibitem [\protect \citeauthoryear {%
Drago%
, Nannicini%
\BCBL {}\ \BBA {} Sobbrio%
}{%
Drago%
\ \protect \BOthers {.}}{%
{\protect \APACyear {2014}}%
}]{%
Dra2014}
\APACinsertmetastar {%
Dra2014}%
\begin{APACrefauthors}%
Drago, F.%
, Nannicini, T.%
\BCBL {}\ \BBA {} Sobbrio, F.%
\end{APACrefauthors}%
\unskip\
\newblock
\APACrefYearMonthDay{2014}{July}{}.
\newblock
{\BBOQ}\APACrefatitle {Meet the Press: How Voters and Politicians Respond to
  Newspaper Entry and Exit} {Meet the press: How voters and politicians respond
  to newspaper entry and exit}.{\BBCQ}
\newblock
\APACjournalVolNumPages{American Economic Journal: Applied
  Economics}{6}{3}{159-88}.
\newblock
\begin{APACrefDOI} \doi{https://doi.org/10.1257/app.6.3.159} \end{APACrefDOI}
\PrintBackRefs{\CurrentBib}

\bibitem [\protect \citeauthoryear {%
Fast%
, Vachovsky%
\BCBL {}\ \BBA {} Bernstein%
}{%
Fast%
\ \protect \BOthers {.}}{%
{\protect \APACyear {2016}}%
}]{%
Fas2016}
\APACinsertmetastar {%
Fas2016}%
\begin{APACrefauthors}%
Fast, E.%
, Vachovsky, T.%
\BCBL {}\ \BBA {} Bernstein, M\BPBI S.%
\end{APACrefauthors}%
\unskip\
\newblock
\APACrefYearMonthDay{2016}{may}{}.
\newblock
{\BBOQ}\APACrefatitle {Shirtless and Dangerous: Quantifying Linguistic Signals
  of Gender Bias in an Online Fiction Writing Community} {Shirtless and
  dangerous: Quantifying linguistic signals of gender bias in an online fiction
  writing community}.{\BBCQ}
\newblock
\BIn{} \APACrefbtitle {Proceedings of the Tenth International AAAI Conference
  on Web and Social Media (ICWSM 2016)} {Proceedings of the tenth international
  aaai conference on web and social media (icwsm 2016)}\ (\BPG~112-120).
\newblock
\APACaddressPublisher{Cologne, Germany}{AAAI Press}.
\PrintBackRefs{\CurrentBib}

\bibitem [\protect \citeauthoryear {%
Gerber%
, Karlan%
\BCBL {}\ \BBA {} Bergan%
}{%
Gerber%
\ \protect \BOthers {.}}{%
{\protect \APACyear {2009}}%
}]{%
Ger2009}
\APACinsertmetastar {%
Ger2009}%
\begin{APACrefauthors}%
Gerber, A\BPBI S.%
, Karlan, D.%
\BCBL {}\ \BBA {} Bergan, D.%
\end{APACrefauthors}%
\unskip\
\newblock
\APACrefYearMonthDay{2009}{April}{}.
\newblock
{\BBOQ}\APACrefatitle {Does the Media Matter? {A} Field Experiment Measuring
  the Effect of Newspapers on Voting Behavior and Political Opinions} {Does the
  media matter? {A} field experiment measuring the effect of newspapers on
  voting behavior and political opinions}.{\BBCQ}
\newblock
\APACjournalVolNumPages{American Economic Journal: Applied
  Economics}{1}{2}{35-52}.
\newblock
\begin{APACrefDOI} \doi{https://doi.org/10.1257/app.1.2.35} \end{APACrefDOI}
\PrintBackRefs{\CurrentBib}

\bibitem [\protect \citeauthoryear {%
Hatzivassiloglou%
\ \BBA {} Wiebe%
}{%
Hatzivassiloglou%
\ \BBA {} Wiebe%
}{%
{\protect \APACyear {2000}}%
}]{%
Hat2000}
\APACinsertmetastar {%
Hat2000}%
\begin{APACrefauthors}%
Hatzivassiloglou, V.%
\BCBT {}\ \BBA {} Wiebe, J\BPBI M.%
\end{APACrefauthors}%
\unskip\
\newblock
\APACrefYearMonthDay{2000}{}{}.
\newblock
{\BBOQ}\APACrefatitle {Effects of Adjective Orientation and Gradability on
  Sentence Subjectivity} {Effects of adjective orientation and gradability on
  sentence subjectivity}.{\BBCQ}
\newblock
\BIn{} \APACrefbtitle {Proceedings of the 18th Conference on Computational
  Linguistics - Volume 1} {Proceedings of the 18th conference on computational
  linguistics - volume 1}\ (\BPG~299–305).
\newblock
\APACaddressPublisher{USA}{Association for Computational Linguistics}.
\newblock
\begin{APACrefDOI} \doi{https://doi.org/10.3115/990820.990864} \end{APACrefDOI}
\PrintBackRefs{\CurrentBib}

\bibitem [\protect \citeauthoryear {%
Honnibal%
, Montani%
, Van~Landeghem%
\BCBL {}\ \BBA {} Boyd%
}{%
Honnibal%
\ \protect \BOthers {.}}{%
{\protect \APACyear {2020}}%
}]{%
Hon2020}
\APACinsertmetastar {%
Hon2020}%
\begin{APACrefauthors}%
Honnibal, M.%
, Montani, I.%
, Van~Landeghem, S.%
\BCBL {}\ \BBA {} Boyd, A.%
\end{APACrefauthors}%
\unskip\
\newblock
\APACrefYearMonthDay{2020}{}{}.
\newblock
\APACrefbtitle {{spaCy: Industrial-strength Natural Language Processing in
  Python}.} {{spaCy: Industrial-strength Natural Language Processing in
  Python}.}
\newblock
\APACaddressPublisher{}{Zenodo}.
\newblock
\begin{APACrefDOI} \doi{https://doi.org/10.5281/zenodo.1212303}
  \end{APACrefDOI}
\PrintBackRefs{\CurrentBib}

\bibitem [\protect \citeauthoryear {%
Koenig%
, Eagly%
, Mitchell%
\BCBL {}\ \BBA {} Ristikari%
}{%
Koenig%
\ \protect \BOthers {.}}{%
{\protect \APACyear {2011}}%
}]{%
Koe2011}
\APACinsertmetastar {%
Koe2011}%
\begin{APACrefauthors}%
Koenig, A\BPBI M.%
, Eagly, A\BPBI H.%
, Mitchell, A\BPBI A.%
\BCBL {}\ \BBA {} Ristikari, T.%
\end{APACrefauthors}%
\unskip\
\newblock
\APACrefYearMonthDay{2011}{}{}.
\newblock
{\BBOQ}\APACrefatitle {Are leader stereotypes masculine? A meta-analysis of
  three research paradigms} {Are leader stereotypes masculine? a meta-analysis
  of three research paradigms}.{\BBCQ}
\newblock
\APACjournalVolNumPages{Psychological Bulletin}{137}{4}{616-642}.
\newblock
\begin{APACrefDOI} \doi{https://doi.org/10.1037/a0023557} \end{APACrefDOI}
\PrintBackRefs{\CurrentBib}

\bibitem [\protect \citeauthoryear {%
Landerer%
}{%
Landerer%
}{%
{\protect \APACyear {2013}}%
}]{%
Lan2013}
\APACinsertmetastar {%
Lan2013}%
\begin{APACrefauthors}%
Landerer, N.%
\end{APACrefauthors}%
\unskip\
\newblock
\APACrefYearMonthDay{2013}{}{}.
\newblock
{\BBOQ}\APACrefatitle {Rethinking the Logics: A Conceptual Framework for the
  Mediatization of Politics} {Rethinking the logics: A conceptual framework for
  the mediatization of politics}.{\BBCQ}
\newblock
\APACjournalVolNumPages{Communication Theory}{23}{3}{239-258}.
\newblock
\begin{APACrefDOI} \doi{https://doi.org/10.1111/comt.12013} \end{APACrefDOI}
\PrintBackRefs{\CurrentBib}

\bibitem [\protect \citeauthoryear {%
Leti%
}{%
Leti%
}{%
{\protect \APACyear {1983}}%
}]{%
Let1983}
\APACinsertmetastar {%
Let1983}%
\begin{APACrefauthors}%
Leti, G.%
\end{APACrefauthors}%
\unskip\
\newblock
\APACrefYear{1983}.
\newblock
\APACrefbtitle {Statistica Descrittiva} {Statistica descrittiva}.
\newblock
\APACaddressPublisher{}{Il Mulino}.
\PrintBackRefs{\CurrentBib}

\bibitem [\protect \citeauthoryear {%
Master%
, Meltzoff%
\BCBL {}\ \BBA {} Cheryan%
}{%
Master%
\ \protect \BOthers {.}}{%
{\protect \APACyear {2021}}%
}]{%
Mas2021}
\APACinsertmetastar {%
Mas2021}%
\begin{APACrefauthors}%
Master, A.%
, Meltzoff, A\BPBI N.%
\BCBL {}\ \BBA {} Cheryan, S.%
\end{APACrefauthors}%
\unskip\
\newblock
\APACrefYearMonthDay{2021}{}{}.
\newblock
{\BBOQ}\APACrefatitle {Gender stereotypes about interests start early and cause
  gender disparities in computer science and engineering} {Gender stereotypes
  about interests start early and cause gender disparities in computer science
  and engineering}.{\BBCQ}
\newblock
\APACjournalVolNumPages{Proceedings of the National Academy of
  Sciences}{118}{48}{}.
\newblock
\begin{APACrefDOI} \doi{https://doi.org/10.1073/pnas.2100030118}
  \end{APACrefDOI}
\PrintBackRefs{\CurrentBib}

\bibitem [\protect \citeauthoryear {%
\"{O}rnebring%
\ \BBA {} J\"{o}nsson%
}{%
\"{O}rnebring%
\ \BBA {} J\"{o}nsson%
}{%
{\protect \APACyear {2004}}%
}]{%
Orn2004}
\APACinsertmetastar {%
Orn2004}%
\begin{APACrefauthors}%
\"{O}rnebring, H.%
\BCBT {}\ \BBA {} J\"{o}nsson, A\BPBI M.%
\end{APACrefauthors}%
\unskip\
\newblock
\APACrefYearMonthDay{2004}{}{}.
\newblock
{\BBOQ}\APACrefatitle {Tabloid journalism and the public sphere: a historical
  perspective on tabloid journalism} {Tabloid journalism and the public sphere:
  a historical perspective on tabloid journalism}.{\BBCQ}
\newblock
\APACjournalVolNumPages{Journalism Studies}{5}{3}{283-295}.
\newblock
\begin{APACrefDOI} \doi{https://doi.org/10.1080/1461670042000246052}
  \end{APACrefDOI}
\PrintBackRefs{\CurrentBib}

\bibitem [\protect \citeauthoryear {%
O’Neill%
, Savigny%
\BCBL {}\ \BBA {} Cann%
}{%
O’Neill%
\ \protect \BOthers {.}}{%
{\protect \APACyear {2016}}%
}]{%
One2016}
\APACinsertmetastar {%
One2016}%
\begin{APACrefauthors}%
O’Neill, D.%
, Savigny, H.%
\BCBL {}\ \BBA {} Cann, V.%
\end{APACrefauthors}%
\unskip\
\newblock
\APACrefYearMonthDay{2016}{}{}.
\newblock
{\BBOQ}\APACrefatitle {Women politicians in the {UK} press: not seen and not
  heard?} {Women politicians in the {UK} press: not seen and not heard?}{\BBCQ}
\newblock
\APACjournalVolNumPages{Feminist Media Studies}{16}{2}{293-307}.
\newblock
\begin{APACrefDOI} \doi{https://doi.org/10.1080/14680777.2015.1092458}
  \end{APACrefDOI}
\PrintBackRefs{\CurrentBib}

\bibitem [\protect \citeauthoryear {%
Pang%
\ \BBA {} Lee%
}{%
Pang%
\ \BBA {} Lee%
}{%
{\protect \APACyear {2008}}%
}]{%
Pan2008}
\APACinsertmetastar {%
Pan2008}%
\begin{APACrefauthors}%
Pang, B.%
\BCBT {}\ \BBA {} Lee, L.%
\end{APACrefauthors}%
\unskip\
\newblock
\APACrefYearMonthDay{2008}{jan}{}.
\newblock
{\BBOQ}\APACrefatitle {Opinion Mining and Sentiment Analysis} {Opinion mining
  and sentiment analysis}.{\BBCQ}
\newblock
\APACjournalVolNumPages{Found. Trends Inf. Retr.}{2}{1–2}{1–135}.
\newblock
\begin{APACrefDOI} \doi{https://doi.org/10.1561/1500000011} \end{APACrefDOI}
\PrintBackRefs{\CurrentBib}

\bibitem [\protect \citeauthoryear {%
Simonton%
}{%
Simonton%
}{%
{\protect \APACyear {1986}}%
}]{%
Sim1986}
\APACinsertmetastar {%
Sim1986}%
\begin{APACrefauthors}%
Simonton, D\BPBI K.%
\end{APACrefauthors}%
\unskip\
\newblock
\APACrefYearMonthDay{1986}{}{}.
\newblock
{\BBOQ}\APACrefatitle {Presidential personality: Biographical use of the
  {G}ough Adjective Check List} {Presidential personality: Biographical use of
  the {G}ough adjective check list}.{\BBCQ}
\newblock
\APACjournalVolNumPages{Journal of Personality and Social
  Psychology}{51}{1}{149–160}.
\newblock
\begin{APACrefDOI} \doi{https://doi.org/10.1037/0022-3514.51.1.149}
  \end{APACrefDOI}
\PrintBackRefs{\CurrentBib}

\bibitem [\protect \citeauthoryear {%
Skovsgaard%
}{%
Skovsgaard%
}{%
{\protect \APACyear {2014}}%
}]{%
Sko2014}
\APACinsertmetastar {%
Sko2014}%
\begin{APACrefauthors}%
Skovsgaard, M.%
\end{APACrefauthors}%
\unskip\
\newblock
\APACrefYearMonthDay{2014}{}{}.
\newblock
{\BBOQ}\APACrefatitle {A tabloid mind? Professional values and organizational
  pressures as explanations of tabloid journalism} {A tabloid mind?
  professional values and organizational pressures as explanations of tabloid
  journalism}.{\BBCQ}
\newblock
\APACjournalVolNumPages{Media, Culture \& Society}{36}{2}{200-218}.
\newblock
\begin{APACrefDOI} \doi{https://doi.org/10.1177/0163443713515740}
  \end{APACrefDOI}
\PrintBackRefs{\CurrentBib}

\bibitem [\protect \citeauthoryear {%
Trimble%
, Wagner%
, Sampert%
, Raphael%
\BCBL {}\ \BBA {} Gerrits%
}{%
Trimble%
\ \protect \BOthers {.}}{%
{\protect \APACyear {2013}}%
}]{%
Tri2013}
\APACinsertmetastar {%
Tri2013}%
\begin{APACrefauthors}%
Trimble, L.%
, Wagner, A.%
, Sampert, S.%
, Raphael, D.%
\BCBL {}\ \BBA {} Gerrits, B.%
\end{APACrefauthors}%
\unskip\
\newblock
\APACrefYearMonthDay{2013}{}{}.
\newblock
{\BBOQ}\APACrefatitle {Is It Personal? {G}endered Mediation in Newspaper
  Coverage of {C}anadian National Party Leadership Contests, 1975–2012} {Is
  it personal? {G}endered mediation in newspaper coverage of {C}anadian
  national party leadership contests, 1975–2012}.{\BBCQ}
\newblock
\APACjournalVolNumPages{The International Journal of
  Press/Politics}{18}{4}{462-481}.
\newblock
\begin{APACrefDOI} \doi{https://doi.org/10.1177/1940161213495455}
  \end{APACrefDOI}
\PrintBackRefs{\CurrentBib}

\bibitem [\protect \citeauthoryear {%
Van~Aelst%
, Sheafer%
\BCBL {}\ \BBA {} Stanyer%
}{%
Van~Aelst%
\ \protect \BOthers {.}}{%
{\protect \APACyear {2012}}%
}]{%
Van2012}
\APACinsertmetastar {%
Van2012}%
\begin{APACrefauthors}%
Van~Aelst, P.%
, Sheafer, T.%
\BCBL {}\ \BBA {} Stanyer, J.%
\end{APACrefauthors}%
\unskip\
\newblock
\APACrefYearMonthDay{2012}{}{}.
\newblock
{\BBOQ}\APACrefatitle {The personalization of mediated political communication:
  A review of concepts, operationalizations and key findings} {The
  personalization of mediated political communication: A review of concepts,
  operationalizations and key findings}.{\BBCQ}
\newblock
\APACjournalVolNumPages{Journalism}{13}{2}{203-220}.
\newblock
\begin{APACrefDOI} \doi{https://doi.org/10.1177/1464884911427802}
  \end{APACrefDOI}
\PrintBackRefs{\CurrentBib}

\bibitem [\protect \citeauthoryear {%
Van~der Pas%
\ \BBA {} Aaldering%
}{%
Van~der Pas%
\ \BBA {} Aaldering%
}{%
{\protect \APACyear {2020}}%
}]{%
Van2020}
\APACinsertmetastar {%
Van2020}%
\begin{APACrefauthors}%
Van~der Pas, D\BPBI J.%
\BCBT {}\ \BBA {} Aaldering, L.%
\end{APACrefauthors}%
\unskip\
\newblock
\APACrefYearMonthDay{2020}{02}{}.
\newblock
{\BBOQ}\APACrefatitle {{Gender Differences in Political Media Coverage: {A}
  Meta-Analysis}} {{Gender Differences in Political Media Coverage: {A}
  Meta-Analysis}}.{\BBCQ}
\newblock
\APACjournalVolNumPages{Journal of Communication}{70}{1}{114-143}.
\newblock
\begin{APACrefDOI} \doi{https://doi.org/10.1093/joc/jqz046} \end{APACrefDOI}
\PrintBackRefs{\CurrentBib}

\bibitem [\protect \citeauthoryear {%
Wagner%
, Trimble%
\BCBL {}\ \BBA {} Sampert%
}{%
Wagner%
\ \protect \BOthers {.}}{%
{\protect \APACyear {2019}}%
}]{%
Wag2019}
\APACinsertmetastar {%
Wag2019}%
\begin{APACrefauthors}%
Wagner, A.%
, Trimble, L.%
\BCBL {}\ \BBA {} Sampert, S.%
\end{APACrefauthors}%
\unskip\
\newblock
\APACrefYearMonthDay{2019}{}{}.
\newblock
{\BBOQ}\APACrefatitle {One Smart Politician: Gendered Media Discourses of
  Political Leadership in {C}anada} {One smart politician: Gendered media
  discourses of political leadership in {C}anada}.{\BBCQ}
\newblock
\APACjournalVolNumPages{Canadian Journal of Political
  Science}{52}{1}{141–162}.
\newblock
\begin{APACrefDOI} \doi{https://doi.org/10.1017/S0008423918000471}
  \end{APACrefDOI}
\PrintBackRefs{\CurrentBib}

\bibitem [\protect \citeauthoryear {%
Whitelaw%
, Garg%
\BCBL {}\ \BBA {} Argamon%
}{%
Whitelaw%
\ \protect \BOthers {.}}{%
{\protect \APACyear {2005}}%
}]{%
Whi2005}
\APACinsertmetastar {%
Whi2005}%
\begin{APACrefauthors}%
Whitelaw, C.%
, Garg, N.%
\BCBL {}\ \BBA {} Argamon, S.%
\end{APACrefauthors}%
\unskip\
\newblock
\APACrefYearMonthDay{2005}{}{}.
\newblock
{\BBOQ}\APACrefatitle {Using Appraisal Groups for Sentiment Analysis} {Using
  appraisal groups for sentiment analysis}.{\BBCQ}
\newblock
\BIn{} \APACrefbtitle {Proceedings of the 14th ACM International Conference on
  Information and Knowledge Management} {Proceedings of the 14th acm
  international conference on information and knowledge management}\
  (\BPG~625–631).
\newblock
\APACaddressPublisher{New York, NY, USA}{Association for Computing Machinery}.
\newblock
\begin{APACrefDOI} \doi{https://doi.org/10.1145/1099554.1099714}
  \end{APACrefDOI}
\PrintBackRefs{\CurrentBib}

\bibitem [\protect \citeauthoryear {%
Wilks%
\ \BBA {} Stevenson%
}{%
Wilks%
\ \BBA {} Stevenson%
}{%
{\protect \APACyear {1998}}%
}]{%
Wil1998}
\APACinsertmetastar {%
Wil1998}%
\begin{APACrefauthors}%
Wilks, Y.%
\BCBT {}\ \BBA {} Stevenson, M.%
\end{APACrefauthors}%
\unskip\
\newblock
\APACrefYearMonthDay{1998}{}{}.
\newblock
{\BBOQ}\APACrefatitle {The grammar of sense: Using part-of-speech tags as a
  first step in semantic disambiguation} {The grammar of sense: Using
  part-of-speech tags as a first step in semantic disambiguation}.{\BBCQ}
\newblock
\APACjournalVolNumPages{Natural Language Engineering}{4}{2}{135–143}.
\newblock
\begin{APACrefDOI} \doi{https://doi.org/10.1017/S1351324998001946}
  \end{APACrefDOI}
\PrintBackRefs{\CurrentBib}

\bibitem [\protect \citeauthoryear {%
Yu%
\ \BBA {} Hatzivassiloglou%
}{%
Yu%
\ \BBA {} Hatzivassiloglou%
}{%
{\protect \APACyear {2003}}%
}]{%
Yu2003}
\APACinsertmetastar {%
Yu2003}%
\begin{APACrefauthors}%
Yu, H.%
\BCBT {}\ \BBA {} Hatzivassiloglou, V.%
\end{APACrefauthors}%
\unskip\
\newblock
\APACrefYearMonthDay{2003}{}{}.
\newblock
{\BBOQ}\APACrefatitle {Towards Answering Opinion Questions: Separating Facts
  from Opinions and Identifying the Polarity of Opinion Sentences} {Towards
  answering opinion questions: Separating facts from opinions and identifying
  the polarity of opinion sentences}.{\BBCQ}
\newblock
\BIn{} \APACrefbtitle {Proceedings of the 2003 Conference on Empirical Methods
  in Natural Language Processing} {Proceedings of the 2003 conference on
  empirical methods in natural language processing}\ (\BPG~129–136).
\newblock
\APACaddressPublisher{USA}{Association for Computational Linguistics}.
\newblock
\begin{APACrefDOI} \doi{https://doi.org/10.3115/1119355.1119372}
  \end{APACrefDOI}
\PrintBackRefs{\CurrentBib}

\end{thebibliography}


\begin{thebibliography}{}

\bibitem [\protect \citeauthoryear {%
Aaldering%
, van~der Meer%
\BCBL {}\ \BBA {} Van~der Brug%
}{%
Aaldering%
\ \protect \BOthers {.}}{%
{\protect \APACyear {2018}}%
}]{%
Aal2018}
\APACinsertmetastar {%
Aal2018}%
\begin{APACrefauthors}%
Aaldering, L.%
, van~der Meer, T.%
\BCBL {}\ \BBA {} Van~der Brug, W.%
\end{APACrefauthors}%
\unskip\
\newblock
\APACrefYearMonthDay{2018}{}{}.
\newblock
{\BBOQ}\APACrefatitle {Mediated Leader Effects: The Impact of Newspapers'
  Portrayal of Party Leadership on Electoral Support} {Mediated leader effects:
  The impact of newspapers' portrayal of party leadership on electoral
  support}.{\BBCQ}
\newblock
\APACjournalVolNumPages{The International Journal of
  Press/Politics}{23}{1}{70-94}.
\newblock
\begin{APACrefDOI} \doi{https://doi.org/10.1177/1940161217740696}
  \end{APACrefDOI}
\PrintBackRefs{\CurrentBib}

\bibitem [\protect \citeauthoryear {%
Aaldering%
\ \BBA {} Vliegenthart%
}{%
Aaldering%
\ \BBA {} Vliegenthart%
}{%
{\protect \APACyear {2016}}%
}]{%
Aal2016}
\APACinsertmetastar {%
Aal2016}%
\begin{APACrefauthors}%
Aaldering, L.%
\BCBT {}\ \BBA {} Vliegenthart, R.%
\end{APACrefauthors}%
\unskip\
\newblock
\APACrefYearMonthDay{2016}{}{}.
\newblock
{\BBOQ}\APACrefatitle {Political leaders and the media. {C}an we measure
  political leadership images in newspapers using computer-assisted content
  analysis?} {Political leaders and the media. {C}an we measure political
  leadership images in newspapers using computer-assisted content
  analysis?}{\BBCQ}
\newblock
\APACjournalVolNumPages{Quality \& Quantity}{50}{5}{1871-1905}.
\newblock
\begin{APACrefDOI} \doi{https://doi.org/10.1007/s11135-015-0242-9}
  \end{APACrefDOI}
\PrintBackRefs{\CurrentBib}

\bibitem [\protect \citeauthoryear {%
Arce%
}{%
Arce%
}{%
{\protect \APACyear {2004}}%
}]{%
Arc2004}
\APACinsertmetastar {%
Arc2004}%
\begin{APACrefauthors}%
Arce, G\BPBI R.%
\end{APACrefauthors}%
\unskip\
\newblock
\APACrefYear{2004}.
\newblock
\APACrefbtitle {Nonlinear Signal Processing: A Statistical Approach} {Nonlinear
  signal processing: A statistical approach}.
\newblock
\APACaddressPublisher{}{Wiley}.
\PrintBackRefs{\CurrentBib}

\bibitem [\protect \citeauthoryear {%
Hollanders%
\ \BBA {} Vliegenthart%
}{%
Hollanders%
\ \BBA {} Vliegenthart%
}{%
{\protect \APACyear {2011}}%
}]{%
Hol2011}
\APACinsertmetastar {%
Hol2011}%
\begin{APACrefauthors}%
Hollanders, D.%
\BCBT {}\ \BBA {} Vliegenthart, R.%
\end{APACrefauthors}%
\unskip\
\newblock
\APACrefYearMonthDay{2011}{}{}.
\newblock
{\BBOQ}\APACrefatitle {The influence of negative newspaper coverage on consumer
  confidence: The {D}utch case} {The influence of negative newspaper coverage
  on consumer confidence: The {D}utch case}.{\BBCQ}
\newblock
\APACjournalVolNumPages{Journal of Economic Psychology}{32}{3}{367-373}.
\newblock
\begin{APACrefDOI} \doi{https://doi.org/10.1016/j.joep.2011.01.003}
  \end{APACrefDOI}
\PrintBackRefs{\CurrentBib}

\bibitem [\protect \citeauthoryear {%
Jeffreys%
\ \BBA {} Jeffreys%
}{%
Jeffreys%
\ \BBA {} Jeffreys%
}{%
{\protect \APACyear {1999}}%
}]{%
Jef1999}
\APACinsertmetastar {%
Jef1999}%
\begin{APACrefauthors}%
Jeffreys, H.%
\BCBT {}\ \BBA {} Jeffreys, B.%
\end{APACrefauthors}%
\unskip\
\newblock
\APACrefYear{1999}.
\newblock
\APACrefbtitle {Methods of Mathematical Physics} {Methods of mathematical
  physics}\ (\PrintOrdinal{3}\ \BEd).
\newblock
\APACaddressPublisher{}{Cambridge University Press}.
\newblock
\begin{APACrefDOI} \doi{https://doi.org/10.1017/CBO9781139168489}
  \end{APACrefDOI}
\PrintBackRefs{\CurrentBib}

\bibitem [\protect \citeauthoryear {%
Krippendorff%
}{%
Krippendorff%
}{%
{\protect \APACyear {2004}}%
}]{%
Kri2004}
\APACinsertmetastar {%
Kri2004}%
\begin{APACrefauthors}%
Krippendorff, K.%
\end{APACrefauthors}%
\unskip\
\newblock
\APACrefYearMonthDay{2004}{}{}.
\newblock
{\BBOQ}\APACrefatitle {Reliability in Content Analysis} {Reliability in content
  analysis}.{\BBCQ}
\newblock
\APACjournalVolNumPages{Human Communication Research}{30}{3}{411-433}.
\newblock
\begin{APACrefDOI} \doi{https://doi.org/10.1111/j.1468-2958.2004.tb00738.x}
  \end{APACrefDOI}
\PrintBackRefs{\CurrentBib}

\end{thebibliography}

\end{document}

\articletype{}% Specify the article type or omit as appropriate

\title{SUPPLEMENTAL MATERIAL FOR\\Gender stereotypes in the mediated personalization of politics: Empirical evidence from a lexical, syntactic and sentiment analysis}

\maketitle

\section*{Supplementary text}
\subsection*{List of news media sources}
Table \ref{supp-tab:source_list} reports the list of all the news media sources that produced the contents analyzed in the paper, divided by type of source (traditional newspapers or online news outlets).

\begin{longtable}{ll}
	\caption{List of sources that produced the political contents analyzed in the paper. The list is sorted alphabetically and each headline is associated with the corresponding type of source, i.e. traditional newspapers or online news outlets.}\label{supp-tab:source_list}\\
	\hline \multicolumn{1}{l}{\textbf{Headline}} & \multicolumn{1}{c}{\textbf{Source set}} \\ \hline 
	\endfirsthead
	
	\multicolumn{2}{c}%
	{{\bfseries \tablename\ \thetable{} -- continued from previous page}} \\
	\hline \multicolumn{1}{l}{\textbf{Headline}} & \multicolumn{1}{c}{\textbf{Source set}} \\ \hline 
	\endhead
	
	\hline \multicolumn{2}{r}{{Continued on next page}} \\ \hline
	\endfoot
	
	\hline \hline
	\endlastfoot
	
	24Emilia&Online news outlets\\
	4 Minuti&Online news outlets\\
	7per24&Online news outlets\\
	Affari Italiani&Online news outlets\\
	Agorà 24&Online news outlets\\
	Agrigento Oggi&Online news outlets\\
	AgrigentoWeb&Online news outlets\\
	AlQamah&Online news outlets\\
	altarimini.it&Online news outlets\\
	Alto Adige&Traditional newspapers\\
	Ancona Today&Online news outlets\\
	AnconaNotizie&Online news outlets\\
	Arezzo Web&Online news outlets\\
	Augusta Online&Online news outlets\\
	Avellino Today&Online news outlets\\
	Avvenire&Traditional newspapers\\
	Bagheria News&Online news outlets\\
	Bari Today&Online news outlets\\
	Basilicata Notizie&Online news outlets\\
	Blasting News&Online news outlets\\
	Blitz Quotidiano&Online news outlets\\
	Blog Beppe Grillo&Online news outlets\\
	Blog Sicilia&Online news outlets\\
	Blogo&Online news outlets\\
	Bologna Today&Online news outlets\\
	Bologna2000&Online news outlets\\
	Brescia Oggi&Traditional newspapers\\
	Brescia Today&Online news outlets\\
	Brindisi Report&Online news outlets\\
	CalNews.it&Online news outlets\\
	Campania Su Web&Online news outlets\\
	CanicattiWeb&Online news outlets\\
	CastelloIncantato&Online news outlets\\
	CastelVetranoSelinunte&Online news outlets\\
	Catania Oggi&Online news outlets\\
	Catania Today&Online news outlets\\
	Catania46&Online news outlets\\
	Catanzaro Informa&Online news outlets\\
	CefaluNews&Online news outlets\\
	Centonove.it&Online news outlets\\
	Cesena Today&Online news outlets\\
	ChartaBianca&Online news outlets\\
	Chieti Today&Online news outlets\\
	Citta della Spezia&Online news outlets\\
	City News&Online news outlets\\
	CoriglianoCalabro&Online news outlets\\
	Corriere Adriatico&Traditional newspapers\\
	Corriere Comunicazioni&Online news outlets\\
	Corriere del Mezzogiorno&Traditional newspapers\\
	Corriere del Trentino&Traditional newspapers\\
	Corriere del Veneto&Traditional newspapers\\
	Corriere dell'Alto Adige&Traditional newspapers\\
	Corriere dell'Umbria&Traditional newspapers\\
	Corriere Della Calabria&Online news outlets\\
	Corriere della Sera&Traditional newspapers\\
	Corriere delle Alpi&Traditional newspapers\\
	Corriere dello Sport&Traditional newspapers\\
	Corriere dello Sport Stadio&Traditional newspapers\\
	Corriere di Bologna&Traditional newspapers\\
	Corriere di Romagna&Traditional newspapers\\
	Corriere di Sciacca&Online news outlets\\
	Corriere Fiorentino&Traditional newspapers\\
	corrierediroma-news.it&Online news outlets\\
	Cronaca Qui&Traditional newspapers\\
	cronacadelveneto.com&Online news outlets\\
	cronacadiverona.com&Online news outlets\\
	Cronache di Caserta&Traditional newspapers\\
	Cronache di Napoli&Traditional newspapers\\
	cronachemaceratesi.it&Online news outlets\\
	Crotone24News&Online news outlets\\
	Dagospia&Online news outlets\\
	Data Sport&Online news outlets\\
	Dg Mag&Online news outlets\\
	Diritto Di Cronaca&Online news outlets\\
	ECNews&Online news outlets\\
	Eco Di Basilicata&Online news outlets\\
	Economia Sicilia&Online news outlets\\
	EconomyUp&Online news outlets\\
	Edicola Di Pinuccio&Online news outlets\\
	emiliaromagnanews.it&Online news outlets\\
	Estense&Online news outlets\\
	FanPage&Online news outlets\\
	ferrara24ore.it&Online news outlets\\
	Firenze Today&Online news outlets\\
	Foggia Today&Online news outlets\\
	Forli 24 Ore&Online news outlets\\
	Forli Today&Online news outlets\\
	Formiche&Online news outlets\\
	Gazzetta del Sud&Traditional newspapers\\
	Gazzetta Dell'Emilia&Online news outlets\\
	Gazzetta di Mantova&Traditional newspapers\\
	Gazzetta di Modena&Traditional newspapers\\
	Gazzetta di Parma&Traditional newspapers\\
	Gazzetta di Reggio&Traditional newspapers\\
	GazzettaJonica&Online news outlets\\
	Genova Today&Online news outlets\\
	Giornale Del Cilento&Online news outlets\\
	Giornale di Brescia&Traditional newspapers\\
	Giornale di Sicilia&Traditional newspapers\\
	Giornale Il Sud&Online news outlets\\
	Giornale L'Ora&Online news outlets\\
	Giornale Nisseno&Online news outlets\\
	GiornaleDiLipari&Online news outlets\\
	Giornalettismo&Online news outlets\\
	GlPress&Online news outlets\\
	Gomarche&Online news outlets\\
	GrandangoloAgrigento&Online news outlets\\
	Hercole&Online news outlets\\
	Huffington Post&Online news outlets\\
	I Giornali di Sicilia&Online news outlets\\
	Il Centro&Traditional newspapers\\
	Il Cittadino&Traditional newspapers\\
	Il Crotonese&Online news outlets\\
	Il Dispaccio&Online news outlets\\
	Il Dubbio&Traditional newspapers\\
	Il Fatto Nisseno&Online news outlets\\
	Il Fatto Quotidiano&Traditional newspapers\\
	Il Fogliettone&Online news outlets\\
	Il Foglio&Traditional newspapers\\
	Il Gazzettino&Traditional newspapers\\
	Il Giornale&Traditional newspapers\\
	Il Giornale D'Italia&Online news outlets\\
	Il Giornale Di Vicenza&Traditional newspapers\\
	Il Giorno&Traditional newspapers\\
	Il Lametino&Online news outlets\\
	Il Manifesto&Traditional newspapers\\
	Il Mattino&Traditional newspapers\\
	Il Mattino di Padova&Traditional newspapers\\
	Il Messaggero&Traditional newspapers\\
	Il Nuovo Giornale di Modena&Online news outlets\\
	Il Pescara&Online news outlets\\
	Il Piacenza&Online news outlets\\
	Il Piccolo&Traditional newspapers\\
	Il Post&Online news outlets\\
	Il Quaderno.it&Online news outlets\\
	Il Resto del Carlino&Traditional newspapers\\
	Il Roma&Traditional newspapers\\
	Il Secolo XIX&Traditional newspapers\\
	Il Sole 24 Ore&Traditional newspapers\\
	Il Tempo&Traditional newspapers\\
	Il Tirreno&Traditional newspapers\\
	IlCaffeGeopolitico&Online news outlets\\
	ilcittadinodimessina.it&Online news outlets\\
	IlDiarioMetropolitano&Online news outlets\\
	IlDolomiti&Online news outlets\\
	IlFattoVesuviano&Online news outlets\\
	IlNordEstQuotidiano&Online news outlets\\
	IlPaeseNuovo&Online news outlets\\
	IlQuotidianoItaliano&Online news outlets\\
	ilsussidiario.net&Online news outlets\\
	Infiltrato&Online news outlets\\
	IonioNotizie&Online news outlets\\
	IrpiniaNews&Online news outlets\\
	Italia Oggi&Traditional newspapers\\
	Key4Biz&Online news outlets\\
	L'Adige&Traditional newspapers\\
	L'Arena&Traditional newspapers\\
	L'Eco di Bergamo&Traditional newspapers\\
	L'Eco di Parma&Online news outlets\\
	L'Osservatore Romano&Traditional newspapers\\
	L'Unione Sarda&Traditional newspapers\\
	L'Unità&Traditional newspapers\\
	La Città di Salerno&Traditional newspapers\\
	La Gazzetta del Mezzogiorno&Traditional newspapers\\
	La Gazzetta dello Sport&Traditional newspapers\\
	La Gazzetta Ennese&Online news outlets\\
	La Gazzetta Trapanese&Online news outlets\\
	La Nazione&Traditional newspapers\\
	La Nota 7&Online news outlets\\
	La Nuova di Venezia e Mestre&Traditional newspapers\\
	La Nuova Ferrara&Traditional newspapers\\
	La Nuova Sardegna&Traditional newspapers\\
	La Prealpina&Traditional newspapers\\
	La Provincia di Como&Traditional newspapers\\
	La Provincia di Cosenza&Traditional newspapers\\
	La Provincia di Lecco&Traditional newspapers\\
	La Provincia di Sondrio&Traditional newspapers\\
	La Provincia di Varese&Traditional newspapers\\
	La Provincia Pavese&Traditional newspapers\\
	La Repubblica&Traditional newspapers\\
	La Riviera Online&Online news outlets\\
	La Sberla&Online news outlets\\
	La Sicilia&Traditional newspapers\\
	La Sicilia Web&Online news outlets\\
	La Stampa&Traditional newspapers\\
	La Tribuna di Treviso&Traditional newspapers\\
	La Verità&Traditional newspapers\\
	La Voce&Online news outlets\\
	La Voce di Mantova&Traditional newspapers\\
	La Voce di Romagna&Traditional newspapers\\
	LAdigetto&Online news outlets\\
	LameziaClick&Online news outlets\\
	lascansione.net&Online news outlets\\
	Latina Quotidiano&Online news outlets\\
	Latina Today&Online news outlets\\
	LaVoceDelNordEst&Online news outlets\\
	LaVoceDelTrentino&Online news outlets\\
	Le Cronache Lucane&Online news outlets\\
	LecceCronaca&Online news outlets\\
	LecceNews24&Online news outlets\\
	LeccePrima&Online news outlets\\
	Lecco Today&Online news outlets\\
	LegnanoNews&Online news outlets\\
	Lettera 43&Online news outlets\\
	Libero&Traditional newspapers\\
	Libero Reporter&Online news outlets\\
	Libertà&Traditional newspapers\\
	lindiscreto.it&Online news outlets\\
	Linkiesta&Online news outlets\\
	lintraprendente.it&Online news outlets\\
	LiveSicilia&Online news outlets\\
	Lo Spiffero&Online news outlets\\
	Lo Strillone&Online news outlets\\
	LOccidentale&Online news outlets\\
	Lugonotizie&Online news outlets\\
	MadonieLive&Online news outlets\\
	Magaze&Online news outlets\\
	Marsala News&Online news outlets\\
	Mazara Online&Online news outlets\\
	Mc Net Tv&Online news outlets\\
	Megachip&Online news outlets\\
	MeridioNews&Online news outlets\\
	Messaggero Veneto&Traditional newspapers\\
	Messina Oggi&Online news outlets\\
	Messina Ora&Online news outlets\\
	MF&Traditional newspapers\\
	Milano Today&Online news outlets\\
	Mo24&Online news outlets\\
	Modena Online&Online news outlets\\
	Modena Today&Online news outlets\\
	Modena2000&Online news outlets\\
	MondoCatania&Online news outlets\\
	Monza Today&Online news outlets\\
	Msn&Online news outlets\\
	Nano Press&Online news outlets\\
	Napoli Today&Online news outlets\\
	Newz&Online news outlets\\
	NordMilano24&Online news outlets\\
	Normanno&Online news outlets\\
	Notizie&Online news outlets\\
	Novara Today&Online news outlets\\
	Nta Calabria&Online news outlets\\
	Nuova Cosenza&Online news outlets\\
	Nuova Società&Online news outlets\\
	Nuovo Sud&Online news outlets\\
	Oggi Milazzo&Online news outlets\\
	Open Online&Online news outlets\\
	Padova News&Online news outlets\\
	Padova Oggi&Online news outlets\\
	Palermo Mania&Online news outlets\\
	Palermo Today&Online news outlets\\
	Parma Online&Online news outlets\\
	Parma Quotidiano&Online news outlets\\
	Parma Today&Online news outlets\\
	ParmaDaily.it&Online news outlets\\
	ParmaReport&Online news outlets\\
	Pavaglione Lugo&Online news outlets\\
	Perugia Today&Online news outlets\\
	Piacenza24&Online news outlets\\
	PiacenzaSera.it&Online news outlets\\
	Picchio News&Online news outlets\\
	Pisa Today&Online news outlets\\
	Piu Notizie&Online news outlets\\
	Puglia Live&Online news outlets\\
	QtSicilia&Online news outlets\\
	Quotidiano di Puglia&Traditional newspapers\\
	Quotidiano di Sicilia&Traditional newspapers\\
	Ragusa Oggi&Online news outlets\\
	RagusaNews&Online news outlets\\
	Ravenna Today&Online news outlets\\
	Ravenna24Ore.it&Online news outlets\\
	RavennaNotizie.it&Online news outlets\\
	ravennawebtv.it&Online news outlets\\
	Redacon&Online news outlets\\
	Reggio Nel Web&Online news outlets\\
	Reggio Report&Online news outlets\\
	Reggio Sera&Online news outlets\\
	Reggio2000&Online news outlets\\
	ResegoneOnline&Online news outlets\\
	Rete News 24&Online news outlets\\
	Rimini Today&Online news outlets\\
	Roma&Traditional newspapers\\
	Roma Today&Online news outlets\\
	Salerno Today&Online news outlets\\
	Sanremo News&Online news outlets\\
	Sardegna Oggi&Online news outlets\\
	Sardinia Post&Online news outlets\\
	Sassari Notizie&Online news outlets\\
	Sassuolo Oggi&Online news outlets\\
	Sassuolo2000&Online news outlets\\
	Savona Notizie&Online news outlets\\
	SciroccoNews&Online news outlets\\
	SecoloTrentino&Online news outlets\\
	SempioneNews&Online news outlets\\
	Settesere&Online news outlets\\
	Si24&Online news outlets\\
	Sicilia Journal&Online news outlets\\
	Sicilia Today&Online news outlets\\
	Sicilia24h&Online news outlets\\
	SiciliaInformazioni&Online news outlets\\
	SiciliaNews24&Online news outlets\\
	Sicilians&Online news outlets\\
	Siracusa Live&Online news outlets\\
	Siracusa News&Online news outlets\\
	Siracusa Oggi&Online news outlets\\
	Stretto Web&Online news outlets\\
	Strill&Online news outlets\\
	SudPress&Online news outlets\\
	TargatoCN&Online news outlets\\
	Telestense&Online news outlets\\
	TempoStretto&Online news outlets\\
	The Social Post&Online news outlets\\
	Tiscali&Online news outlets\\
	Today&Online news outlets\\
	Torino Today&Online news outlets\\
	TP24&Online news outlets\\
	TPI News&Online news outlets\\
	Trapani Oggi&Online news outlets\\
	TrapaniOk&Online news outlets\\
	Trentino&Traditional newspapers\\
	Trento Today&Online news outlets\\
	Treviso Today&Online news outlets\\
	Trieste Prima&Online news outlets\\
	TuttoSport&Traditional newspapers\\
	Udine Today&Online news outlets\\
	Urban Post&Online news outlets\\
	Vai Taormina&Online news outlets\\
	Varese News&Online news outlets\\
	Venezia Today&Online news outlets\\
	Verona Sera&Online news outlets\\
	Vicenza Today&Online news outlets\\
	vivereancona.it&Online news outlets\\
	ViviEnna&Online news outlets\\
	voceditalia.it&Online news outlets\\
	Web Marte&Online news outlets\\
	Yahoo Notizie&Online news outlets\\
	Zoom Sud&Online news outlets\\
\end{longtable}

\subsection*{Preprocessing procedure}
The space of our investigation is represented by the universe of all the articles reported in all national (and multiregional) newspapers and online news outlets during the period from January 2017 to November 2020. The news items collected are filtered according to the occurrence of named entities referring to the political offices under scrutiny. We consider as named entity one of the following mentions:
\begin{itemize}
	\item name + surname, e.g. \textit{Chiara Appendino, Attilio Fontana}
	\item role + surname, e.g. \textit{Governor De Luca, Minister Fedeli, Undersecretary Castelli}
	\item specific role, e.g. \textit{Governor} (or \textit{President}) \textit{of Lazio, Governor} (or \textit{President}) \textit{of the Lazio Region, Mayor of Rome, Minister of Interior}
\end{itemize}
We first perform a sequence of actions to the texts of the resulting collection of news items $\mathcal{D}_c$. These steps include the splitting of contents into sentences and the pruning of sentences not mentioning the entities investigated, the part-of-speech (POS) tagging and the dependency parsing tasks. In addition, the single words are reduced to their base (or lemma) forms by means of a manually created list token-lemma available at \href{https://github.com/brema76/lemmatization-ita}{https://github.com/brema76/lemmatization-ita}. Further, a list of terms which do not add much meaning to a sentence (stopwords) is filtered out together with digits, special characters and url addresses. Second, we exploit the syntactic structures of the remaining sentences in order to select only the words which are more likely to be attributed to the named entity mentioned. In addition, for each of these terms we determine its semantic orientation in the political domain.

\subsection*{Syntactic $n$-grams Vs linear $n$-grams}
\begin{figure}[h]
	\centering
	\includegraphics[width=\columnwidth]{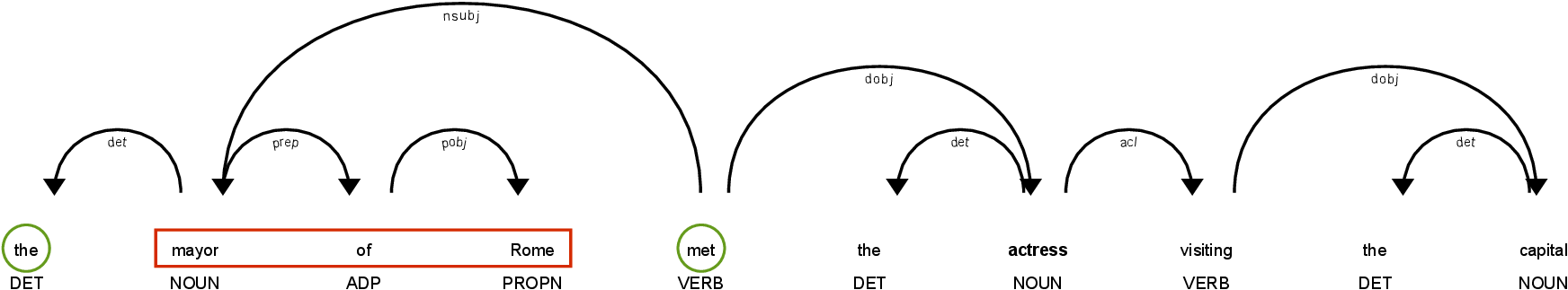}
	\caption{The dependency tree of the example sentence.}\label{supp-fig:tree}
\end{figure}

The personalization literature which relies on computer-assisted content analysis mainly consists in searching for media contents that contain at least one of the words of a pre-specified lexicon within a certain linear distance to the politician under scrutiny \citep{Aal2018,Aal2016,Hol2011}. Nevertheless, for identifying the words in a sentence which are actually attributed to a given target, linear $n$-grams in the sense of adjacent strings of tokens, parts of speech, etc. could be not satisfactory. For instance, consider the sentence
\begin{center}
	\textit{The mayor of Rome met the actress visiting the capital.}	
\end{center}
Since the personalizing word actress is at distance 3 from the named entity mayor of Rome, any system based on linear n-grams with $n\geq 3$ would regard it as referred to the political office holder. One possible approach for overcoming this problem is the use of syntactic $n$-grams. Instead of following the sequential order in the sentence, the linguistic pattern of the words is based on their respective position in the syntactic parse tree. We argue that the words which appear nearby a named entity in the dependency tree are more likely candidates for personalizing expressions. For instance, adjectives generally appear in a dependency tree close to the nouns they describe. Hence, we limit our scope to the syntactic neighborhoods of the named entities which refer to the politicians under scrutiny, by keeping only adjectives, nouns and verbs (except auxiliary and modal). Figure \ref{supp-fig:tree} shows the dependency tree of the aforementioned example sentence.

The words within the red rectangle represent the named entity under scrutiny, whereas the green circles represent the corresponding syntactic neighbors. Since these latter are both excluded from the analysis (\textit{the} is a stopword and \textit{meet} is not in our lexicon), the sentence is pruned, notwithstanding the simultaneous presence of a named entity under investigation (\textit{mayor of Rome}) and a personalizing word (\textit{actor}).

\subsection*{Words as coded units to analyze}
\begin{figure}[h]
	\centering
	\includegraphics[width=\columnwidth]{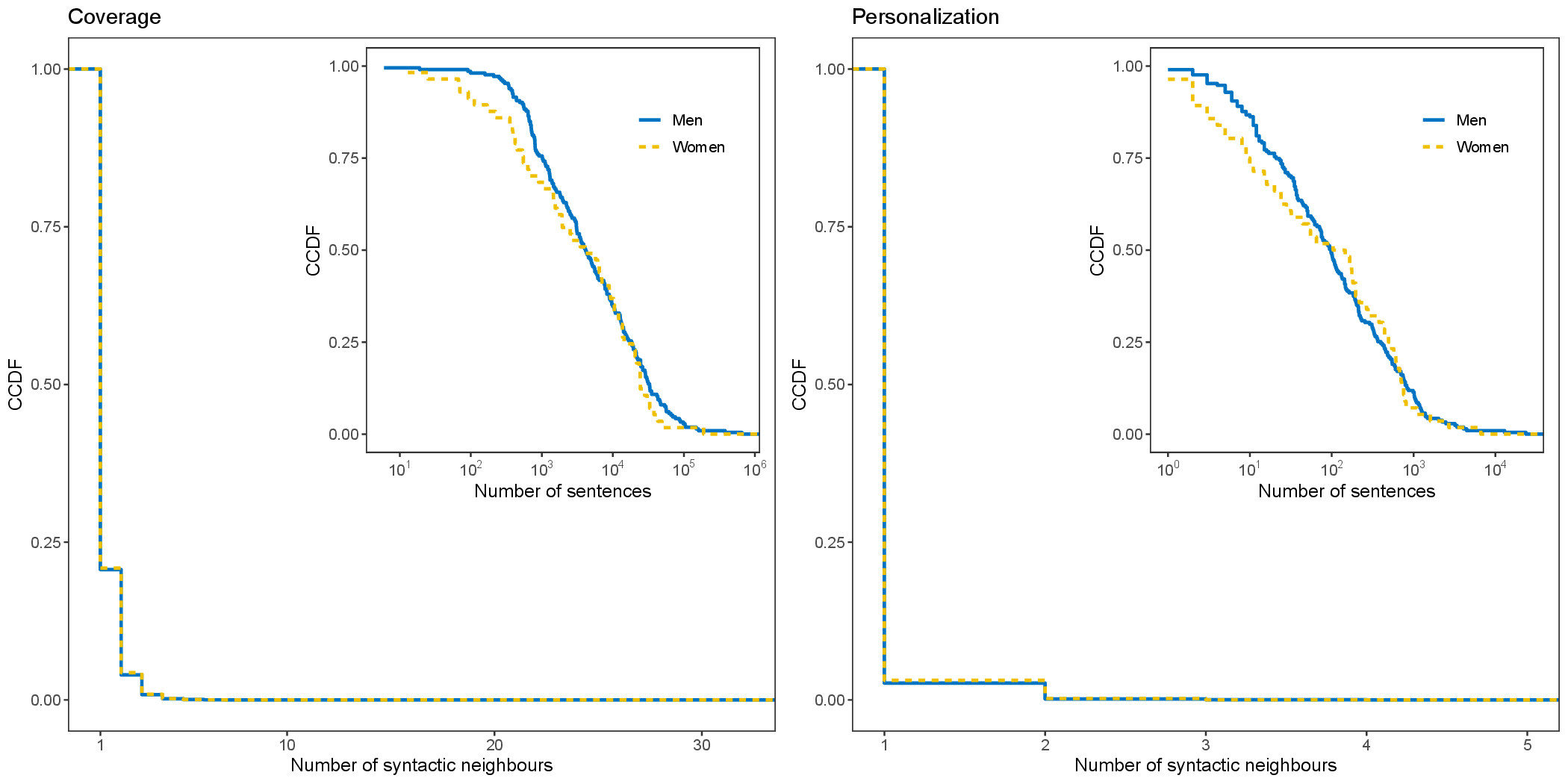}
	\caption{Complementary cumulative distribution function (CCDF) of both the number of syntactic neighbors per sentence (main plots) and the number of sentences per politician (inset plots). Data are divided by both dataset and gender.}\label{supp-fig:ccdf}
\end{figure}

Fig. \ref{supp-fig:ccdf} shows the complementary cumulative distribution function (CCDF) of both the number of syntactic neighbors per sentence (main plots) and the number of sentences per politician (inset plots). Data are divided by both dataset (coverage and personalization, respectively) and gender.

Despite the considerable difference in coverage between women and men, the two representatives exhibit very similar patterns with respect to the number of both sentences and syntactic neighbors. Furthermore, in almost all the sentences in the personalization dataset $\mathcal{D}_p$, the syntactic neighborhood of the named entity mentioned consists of a single lexicon word. Hence, albeit we aim to refer to sentences as coded units to analyze, for the sake of simplicity we consider the single lexicon words instead.

\subsection*{The sentiment classification of personalizing words in the political domain}
The annotators hired for manually assigning a sentiment score to each personalizing word (-1, 0 and 1 for negative, neutral and positive meanings, respectively), are instructed to proceed by contextualizing the words to annotate in the political domain. The task of identifying the semantic orientation of the single words as referrer to political offices requires indeed a particular attention. For instance, the terms \textit{teenager, fairy, powerful, tempting} could have a positive or neutral sentiment in a more general context, but they certainly gain a negative sense when attributed to politicians. To summarize the semantic orientation of a single word in our lexicon, we assign it the average value of the five scores received during the annotation process. According to their aggregate sentiment scores, we further classify the lexicon words as depicted in Table \ref{supp-tab:sentiment_score}.
\begin{table}
	\centering
	\begin{tabular}{rl}
		\toprule
		\textbf{Sentiment} & \textbf{Score values} \\
		\midrule
		strong positive & $\{0.8,1\}$\\
		weakly positive & $\{0.4,0.6\}$\\
		neutral & $\{-0.2,0,0.2\}$\\
		weakly negative & $\{-0.6,-0.4\}$\\
		strong negative & $\{-1,-0.8\}$\\
		\bottomrule
	\end{tabular}
	\caption{Sentiment classification of the lexicon words.}
	\label{supp-tab:sentiment_score}
\end{table}

Aside from the aggregate sentiment score of each lexicon word, we also measure the agreement among annotators as results from the Krippendorff's alpha ($\alpha$). This coefficient accounts for the reliability of the annotation process by returning a real value between 0 (total disagreement) and 1 (perfect agreement). Note that $\alpha$ also accounts for different metrics. Since the sentiment scores assigned by each annotator have the meaning of ranks, we use the ordinal metric \citep{Kri2004}.

\subsection*{The definition of the coverage bias index $I$}
For a word $w$ observed in the coverage dataset $\mathcal{D}_c$, let $|w_F|$ and $|w_M|$ be the counts for women and men, respectively. Let $|F|$ and $|M|$ be the total number of women and men politicians for which at least one record is found in $\mathcal{D}_c$. Let $|D_F|$ and $|D_M|$ be the total number of words addressed to women and men, respectively, so that $|D_T| =|D_F|+|D_M|$ is the total number of words listed in $\mathcal{D}_c$.
Thus, $a_F = \dfrac{|D_F|}{|F|}$ and $a_M = \dfrac{|D_M|}{|M|}$ are the average numbers of words per woman and man, respectively. Given the above notation, consider the incidence rates:
\begin{equation}
	\label{rates0}
	t_F(w) = \dfrac{|w_F|}{|D_F|}, \qquad t_M(w) = \dfrac{|w_M|}{|D_M|}
\end{equation}
reporting the importance of a word count relative to total number of words per women and men, respectively. Given the structural under-presence of women in politics, it is reasonable to find $|D_F|<|D_M|$ and $|F|<|M|$. However, if the average number of words per individual is constant given gender ($a_F \approx a_M$), one could claim that news coverage is homogeneous and women and men are equally represented. In order to adjust the observed incidence rates for gender bias given by \eqref{rates0}, we define the coverage factors to be the importance of $a_F$ and $a_M$ relative to their average $\widetilde{a} = \dfrac{1}{2}\big(a_F + a_M \big)$. Specifically, the proposal is to adjust the total counts $|D_F|$ and $|D_M|$ with correction factors $c_F$ and $c_M$ defined as:
\begin{equation}
	\label{cf}
	c_F = \dfrac{a_F}{\widetilde{a}}, \qquad c_M=\dfrac{a_M}{\widetilde{a}}.
\end{equation}
Consequently, we propose to measure gender bias in coverage in terms of the adjusted incidence rates:
\begin{equation}
	\label{rates1}
	\tilde{t}_F(w) = \dfrac{t_F(w)}{c_F |D_F|}, \qquad \tilde{t}_M(w) = \dfrac{t_M(w)}{c_M |D_M|}
\end{equation}
Clearly, if news coverage is gender-balanced, then both $c_F$ and $c_M$ will be close to 1 and one recovers \eqref{rates0}  from \eqref{rates1}. The smaller $a_F$ is relative to $\tilde{a}$ instead, the stronger is the magnification effect on words' count needed to compare words' incidence rates for women with those of men in order to account for unbalanced coverage. Dually, the larger is $a_M$ with respect to $\tilde{a}$, the higher $c_M$ will and thus the corresponding word's incidence $\tilde{t}_M$ will be more mitigated. 

\subsection*{The reliability of $I$}
\begin{figure}[h]
	\centering
	\includegraphics[width=0.5\columnwidth]{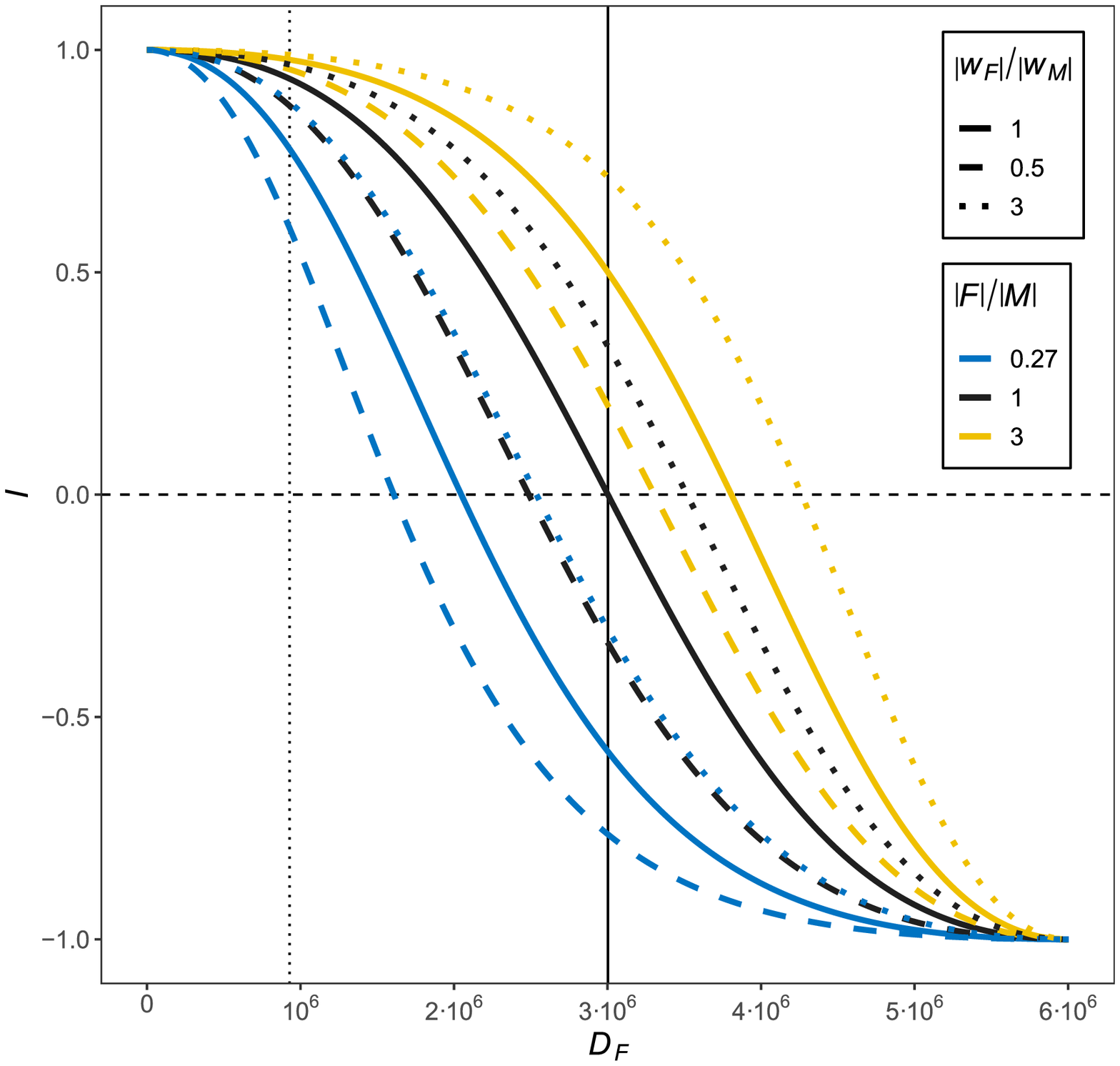}
	\caption{The behavior of the coverage bias index $I$ for different scenarios.}\label{supp-fig:behavior}
\end{figure}

We give an assessment on the reliability of the coverage bias index $I$, by investigating its behavior under different scenarios.
Figure \ref{supp-fig:behavior} shows the values of $I$ (y-axis) for increasing values of $|D_F|$ (x-axis), ranging from $0$ to the observed total number of word counts $|D_T|$ in the coverage dataset.

Let us consider the case of a word $w$ such that $|w_F|=|w_M|$ which corresponds to the solid lines. First focus on the black solid line, corresponding to the circumstance of a sample balanced for gender ($|F|=|M|$). In this case:
\begin{itemize}
	\item $I(w) =0$ if and only if $|D_F| = |D_M| = \frac{|D_T|}{2}$.
	\item If $|D_F| < \frac{|D_T|}{2}$ instead, since $|F|=|M|$, average of words count per women is lower than average words count per men. Thus, $I(w)>0$ for each word $w$ such that $|w_F|=|w_M|$ and the usage of $w$ is positively biased for women (the observed value for $|D_F|$ is identified, for reference, by the vertical dotted line).
\end{itemize}
Then consider the blue solid line, corresponding to the observed unbalanced sample.
\begin{itemize}
	\item Under the scenario $|D_F|=|D_M|$, the density of words per women is higher than it is for men. Thus, $I(w)<0$ for each word $w$ such that $|w_F| = |w_M|$ and $w$ is relatively more used for men than for women.
	\item For a word $w$ such that $|w_F| = |w_M|$, it is possible to find $I(w)=0$ only if $|D_F| < |D_M|$ (see the intersection point of the blue solid line with the line $I=0$). Then, for the unbalanced sample of individuals, for a word $w$ such that $|w_F| = |w_M|$, homogeneity of coverage given gender ($I(w)=0$) is reached only for $|D_F| < |D_M|$. 
	\item The blue solid line is constantly below the black solid line: this indicates that, for all values of $|D_F|$, the coverage bias index $I(w)$ of a word $w$ such that $|w_F| = |w_M|$ is constantly lower if $|F|<|M|$ than if $|F| =|M|$. Indeed, for a fixed value of $|D_F|$, the average number of words per individual is lower for women than it is for men if $|F|<|M|$ than if $|F|=|M|$. 
\end{itemize}
The yellow solid line corresponds to the scenario in which $|F|>|M|$: in particular we set $|F|=3|M|$. In this case, if $|D_F| = |D_M| = \frac{|D_T|}{2}$, for each word $w$ such that $|w_F|=|w_M|$, the coverage index will assume positive values to account for the lower coverage per individual observed for women. 

Lastly, we study the behavior of $I$ for a word $w$ such that $|w_F|\neq |w_M|$. Consider first the case $|w_F| < |w_M|$ (dashed lines of Figure \ref{supp-fig:behavior}), it holds that:
\begin{itemize}
	\item With respect to the case $|w_F|=|w_M|$, in the scenario of equilibrium $|F|=|M|$ and $|D_F|=|D_M|$ (black dashed line), the index value $I(w)$ is lower than 0, correctly reporting that the word is more used for men than it is for women. 
	\item Given this benchmark, if $|F|<|M|$ but $|D_F| = |D_M|$ (blue dashed line), the index value would further decrease to account also for the lower density of words per men with respect to that of women. This circumstance applies for all values of $|D_F|$.
	\item Given the equilibrium benchmark, if $|F|>|M|$ but $|D_F| = |D_M|$ (yellow dashed line), $i$ would increase instead to account for the higher density of words per men with respect to that of women (by penalized the word frequency). This circumstance applies for all values of $|D_F|$.
\end{itemize}
Dual reasoning applies for a word $w$ such that $|w_F|> |w_M|$ (dotted lines of Figure \ref{supp-fig:behavior}).

\subsection*{Comparing the levels of personalized coverage of women and men representatives}
\begin{figure}[h]
	\centering
	\includegraphics[width=0.8\columnwidth]{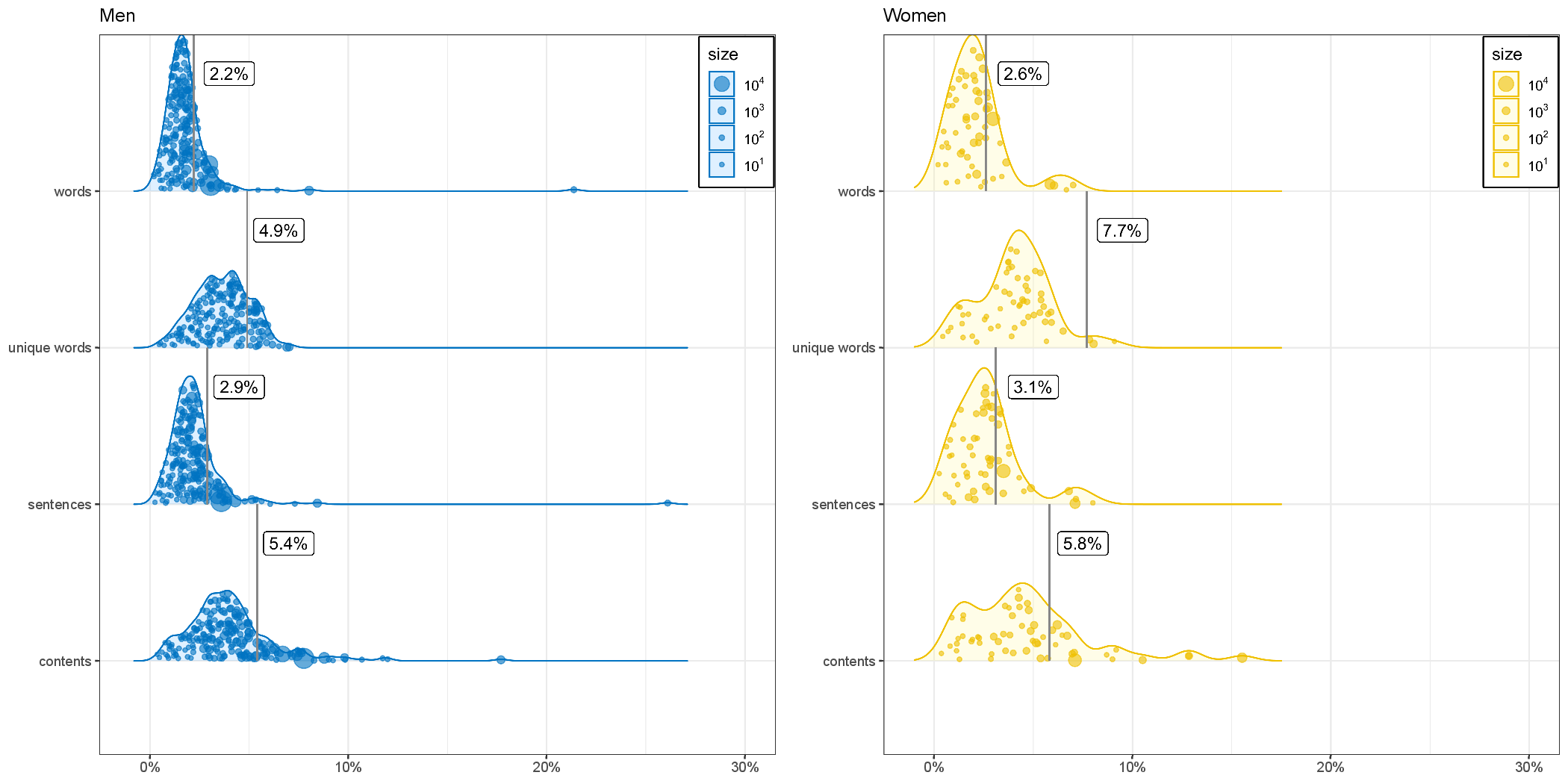}
	\caption{Personalization coverage with respect to different textual units.}\label{supp-fig:PDFs}
\end{figure}

Fig. \ref{supp-fig:PDFs} shows the percentage of media coverage containing references to personal details of the political offices under scrutiny, with respect to different textual units.

Namely, we consider the media contents and the sentences contained therein where a politician is mentioned, as well as the (distinct) personalizing words which constitute the syntactic neighborhood of the corresponding named entity. The observations underlying each empirical Probability Density Function (PDF) curve represent the single politicians and the corresponding sizes the amount of personal coverage they received. The vertical lines indicate the average rates as a result of considering all the politicians as a whole. Except a few offices who, nonetheless, do not attract a significant personal reporting, the fraction of personalized coverage is always below 10\% for every other politician. The breakdown by gender reveals instead that women representative attracts more personal reporting with respect to all the textual units considered, especially words and distinct words.

\begin{figure}[h]
	\centering
	\includegraphics[width=0.7\columnwidth]{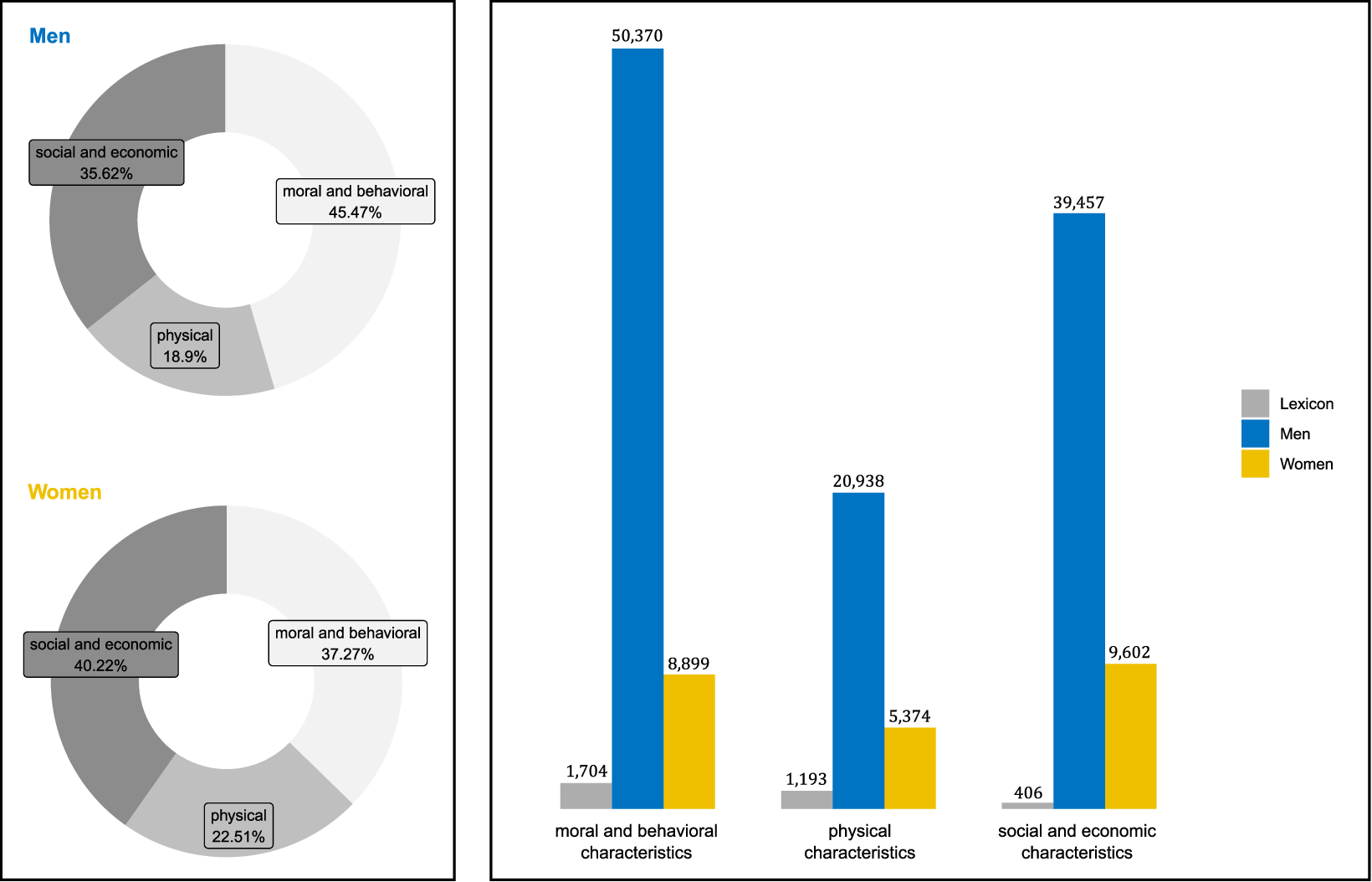}
	\caption{The distribution of the lexicon words between the three facets of the personalization, and the corresponding cumulative frequency with regard to the women and men coverage, respectively (right panel). The fraction of women and men personalized coverage, respectively, with respect to the same facets of the phenomenon (left panel). behavior of the coverage bias index $I$ for different scenarios.}\label{supp-fig:barplots}
\end{figure}

To gain a deeper insight into the nature of this gender-differentiated coverage, we analyze different aspects of the personalization in relation to news content. Namely, each personalized element is classified according to whether it refers to moral-behavioral characteristics, physical characteristics, or socio-economic characteristics. Fig. \ref{supp-fig:barplots} displays the distribution of the lexicon words among the specified categories and the number of times they are used as references to women and men politicians, respectively (right panel). Moreover, it shows that media attention on personal details of women politicians is distributed over the three categories similarly to the men representative (left panel).

Nevertheless, women politicians generally receive more attention on their private life (nearly 2.5\% and 5\% more than their men colleagues with respect to physical and socio-economic characteristics, respectively). To the contrary, there is a greater focus on moral and behavioral characteristics of men politicians (nearly 8\% more than the women representative).

\subsection*{Analysis of the personalization phenomenon through time}
To check whether the observed gaps are due to specific and time-limited events or they reflect the persistence of entrenched gender stereotypes, we also investigate the personalization phenomenon through time. Namely, we consider the time series whose data-points are represented by the daily fraction of personalized coverage concerning each analyzed category for women and men politicians, respectively, and we estimate the underlying trends in each time series by taking a 3-months simple moving average \citep{Arc2004}, i.e. the data-point at time $t$ is given by the mean over the last 90 data-points:
\begin{equation}
	\label{ma_eq}
	\bar{p}(t) = \frac{1}{90}\displaystyle\sum_{\tau=1}^{90}p(t-\tau+1)
\end{equation}
where $p(t)$ is the actual daily fraction of personalized coverage at time $t$.

Coverage on both physical and socio-economic characteristics results almost continuously higher for women than men politicians (89-11\% and 82-18\% of data-points, respectively), suggesting that such personal details typically receive more focus when concerning the women representative. Instead, with respect to moral and behavioral characteristics, neither of the two series is constantly over the other and a number of alternations occurs throughout the period (women prevalence covers 54\% of data-points, men prevalence 46\%).

In addition, to measure the cumulative difference between the two trends, we rely on the area of the region $R$ between the moving average curves $\bar{p}_F(t)$ and $\bar{p}_M(t)$, and bounded on the left and right by the lines $t=t_s$ and $t=t_f$, respectively, where $t_s$ and $t_f$ are the extreme values of the time domain. The area of $R$ is given by
\begin{equation}\label{eq:integration}
	A=\int_{t_s}^{t_f}|\bar{p}_F(t)-\bar{p}_M(t)|dt    
\end{equation}
and it holds $A=A_F+A_M$, where $A_F$ is the area of the region where $\bar{p}_F(t)>\bar{p}_M(t)$ and $A_M$ is the area of the region where $\bar{p}_M(t)>\bar{p}_F(t)$.

Table \ref{supp-tab:simpson} shows the values of $A_F,A_M$ and $A$ for each analyzed category, as a result of the numerical approximation of \eqref{eq:integration} using Simpson's rule \citep{Jef1999}.
\begin{table}[h]
	\centering
	\begin{tabular}{rccc}
		\toprule
		& $A_F$ & $A_M$ & $A$\\
		\midrule
		\textbf{Moral and behavioral}  & 1.073 &  0.996 & 2.069\\
		\textbf{Physical} & 1.770 & 0.044 & 1.814\\
		\textbf{Social and economic} & 4.571 & 0.701 & 5.272 \\
		\bottomrule
	\end{tabular}
	\caption{Area of the region between the moving average curves $\bar{p}_F(t)$ and $\bar{p}_M(t)$, and bounded on the left and right by the lines $t=t_s$ and $t=t_f$, respectively, where $t_s$ and $t_f$ are the extreme values of the time domain. For each analyzed category, $A_F$ is calculated for any $t$ such that $\bar{p}_F(t)>\bar{p}_M(t)$, $A_M$ for any $t$ such that $\bar{p}_M(t)>\bar{p}_F(t)$, and $A$ over the entire time domain.}\label{supp-tab:simpson}
\end{table}

Despite the physical trends define the smallest region, the breakdown by gender reveals the irrelevance of the few parts with a men prevalence. The moving average curves concerning moral and behavioral characteristics limit a little bit larger region, but the parts with women and men prevalence, respectively, are approximately equivalent.
Finally, the socio-economic moving averages are combined with both the biggest region and the greatest difference between areas of the subregions with women and men prevalence, respectively.
Summarizing, the coverage gaps concerning private life (physical appearance and socio-economic background) can be reasonably framed as a result of the persistence of entrenched female stereotypes, being such personal descriptions almost continuously higher for female than male politicians throughout the period. Instead, the lack of a clear dominant trend regarding moral and behavioral characteristics suggests a more mitigated (or at least a more balanced) effect of gender stereotypes.

\subsection*{Gender differences in the content of media coverage}
\begin{figure}[h]
	\centering
	\includegraphics[width=\textwidth]{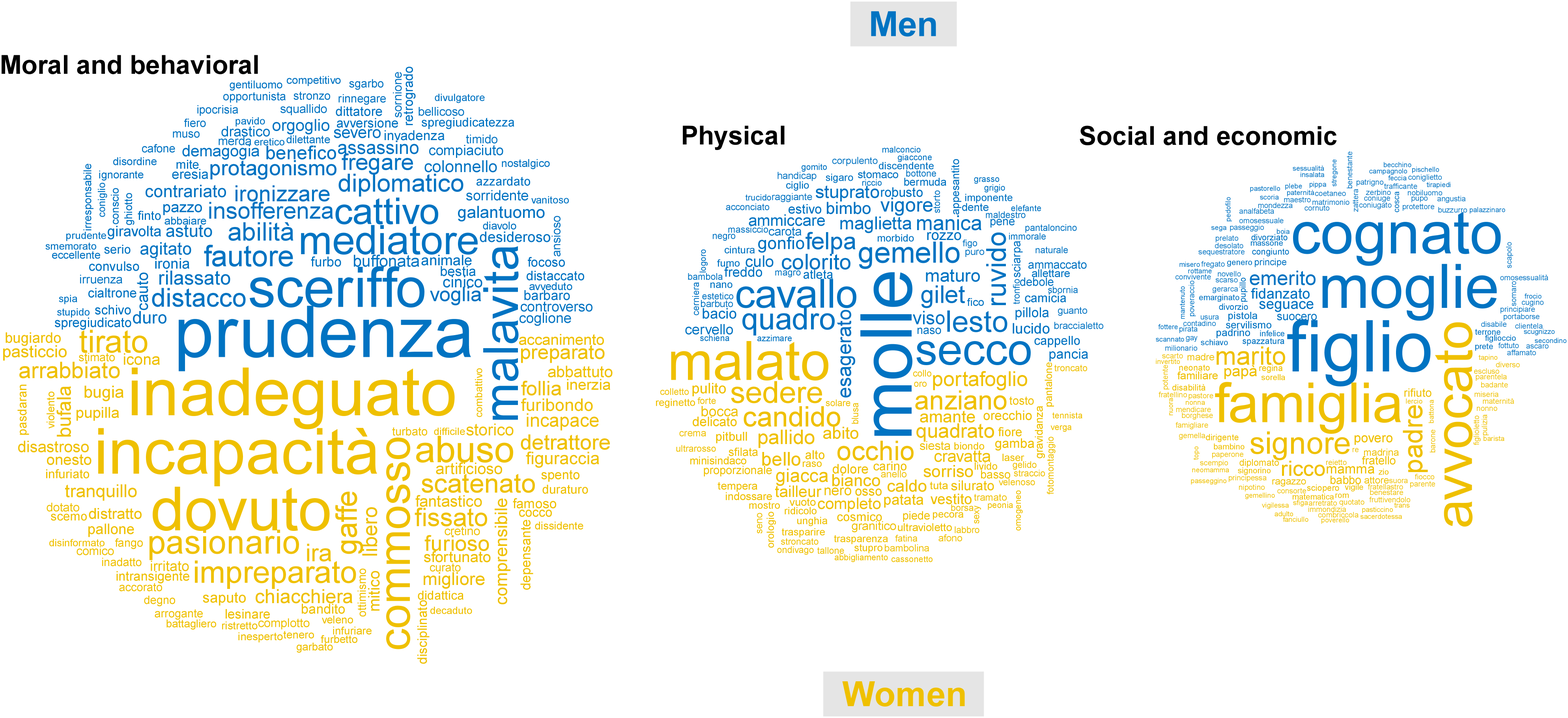}
	\caption{Comparison of the most distinctive personalized words of women and men politicians.}
	\label{supp-fig:wordclouds_freq}
\end{figure}

The wordclouds of Figure \ref{supp-fig:wordclouds_freq} show a comparison of the most distinctive words of women and men politicians, respectively, with regard to each analyzed facet of personalization. A word $w^{\ast}$ belonging to one of the analyzed facet of personalization for which $Diss_{-w^{\ast}} < Diss$ is considered men-distinctive if $\tilde{t}_M(w^{\ast})>\tilde{t}_F(w^{\ast})$, women-distinctive otherwise. The font size of $w^{\ast}$ is proportional to the difference $Diss-Diss_{(-w^{\ast})}$ and represents the dissimilarity of the frequency distributions obtained after omitting $w^{\ast}$ from the dataset.

The Moral and behavioral wordle shows that stereotypically men politicians are depicted as:
\begin{itemize}
	\item powerful: \textit{sheriff/sceriffo, strong/duro, colonel/colonnello, intrusiveness/invadenza, impetuosity/irruenza};
	\item active: \textit{mediator/mediatore, advocate/fautore, ability/abilità, desire/voglia, cynical/cinico, unscrupulous/spregiudicato, fiery/fiero, bold/azzardato, convulsive/convulso, competitive/competitivo};
	\item violent: \textit{underworld/malavita, bad/cattivo, assassin/assassino, beast/bestia, animal/animale, barbarian/barbaro, dictator/dittatore, warlike/bellicoso}.
\end{itemize}
On the contrary, it is impressive how strongly women are perceived as not fit to hold public office: \textit{unfit/inadeguato, incapability/incapacità, unprepared/impreparato, gaffe, madness/follia, incompetent/incapace, unsuitable/inadatto, embarrassment/figuraccia, stupid/scemo, disastrous/disastroso, trouble/pasticcio, jerk/cretino, misinformed/disinformato, inexperienced/inesperto}.

Along this path, concerning social and economic characteristics, it is interesting to note that all the words referred to parenting (\textit{mum/mamma, mother/madre, father/padre, dad/papà-babbo}) are unbalanced towards women, as if to stress the role played by powerful parents in the political careers of their daughters.

With respect to physical characteristics, it is worth to differentiate between physical appearance, clothing and body parts.
With reference to physical appearance, men politicians are mainly depicted with reference to size: \textit{soft/molle, slender/secco, puffy/gonfio, exaggerated/esagerato, robust/robusto, dwarf/nano, imposing/imponente, massive/massiccio, clumsy/maldestro, portly/corpulento, smug/tronfio, fat/grasso, skinny/magro}. On the other hand, women politicians receive a greater deal of focus on their attractiveness: \textit{pretty/bello, smile/sorriso, lover/amante, tall/alto, fashion parade/sfilata, cute/carino, beauty queen/reginetta, baby girl/bambolina, fairy/fatina, sexy}.
With reference to clothing, male politicians are mostly portrayed with casual outfits (\textit{sweatshirt/felpa, vest/gilet, shirt/maglietta, hat/cappello, shorts/pantaloncino, jacket/giaccone}), whereas female with stylish ones (\textit{blouse/blusa, pantsuit/pantalone, dress/abito, suit/completo, blazer/giacca, tailleur, collar/colletto, tie/cravatta}).
Finally, with a few exception in favour of men (\textit{nose/naso, tummy/pancia, stomach/stomaco, back/schiena}), body parts are mentioned more as reference to women (\textit{eye/occhio, backside/sedere, mouth/bocca, ear/orecchio, neck/collo, foot/piede, leg/gamba, bosom/seno, lip/labbro, nail/unghia, blonde hair/biondo}).

The restriction to negative meanings does not produce significant differences with the general wordles of Figure \ref{supp-fig:wordclouds_freq}, as regards to both Moral and behavioral category and Physical category. This implies that most of the gender-distinctive words in such categories are assigned with a negative sentiment.
With reference to socio-economic characteristics, a negative sentiment towards men is mostly associated with underworld and criminal organizations (\textit{adept/seguace, servility/servilismo, gun/pistola, freemason/massone, freemasonry/massoneria, hierarch/gerarca, clique/cosca, gang/cricca, rabble/gentaglia, henchman/tirapiedi, whoremonger/protettore, pimp/pappone, kidnapper/sequestratore, usury/usura, clientelist/clientelare, dealer/trafficante}). On the other hand, a negative sentiment towards women is mainly used to describe their economic status (\textit{rich/ricco, poor/povero, billionaire/miliardario, burgeois/borghese, poverty/miseria, scrooge/paperone, baron/barone, homeless/senzatetto, pauper/meschino, needy/poverello, viscount/visconte}).

\begin{table}[h]
	\centering
	\begin{tabular}{rrrrr}
		\toprule
		& \multicolumn{2}{r}{\textbf{Coverage dataset}} & \multicolumn{2}{r}{\textbf{Personalization dataset}} \\
		& F & M & F & M\\
		\midrule
		\multirow{2}{*}{Traditional newspapers} & 550,681 & 3,106,012 & 14,803 & 71,415\\
		& \tiny\textit{(565,822)} & \tiny\textit{(3,090,871)} & \tiny\textit{(15,289)} & \tiny\textit{(70,929)}\\ 
		\multirow{2}{*}{Online news outlets} & 378,479 & 1,969,639 & 9,072 & 39,350\\
		& \tiny\textit{(363,338)} & \tiny\textit{(1,984,780)} & \tiny\textit{(8,586)} & \tiny\textit{(39,836)}\\ 
		\midrule
		$\chi^2$ statistics & \multicolumn{2}{c}{1225.7} & \multicolumn{2}{c}{52.0}\\
		\bottomrule
	\end{tabular}
	\caption{Words count per gender conditional to both dataset (coverage and personalization) and source type (traditional newspapers and online news outlets). Corresponding $\chi^2$ statistics is reported. Expected frequency under the assumption of independence of coverage between gender of the politician and source type are reported in smaller italics font for each cell.}
	\label{supp-tab:chi2}
\end{table}

\subsection*{Dataset S1 (separate file)}
List of 3,303 personalizing words annotated with the corresponding sentiment classification as referred to political offices. Words are group by category: Moral and behavioral, Physical, Social and economic. 

\clearpage

\bibliographystyle{apacite}
\bibliography{interactapasample_si}